\documentclass[12pt]{article}

\RequirePackage[OT1]{fontenc}
\usepackage{amsmath}
\usepackage{amsmath}
\usepackage{amsfonts}
\usepackage{amssymb}
\usepackage{amsthm}   
\usepackage{bbm}
\usepackage{empheq}   
\usepackage{enumerate}
\usepackage{float}
\usepackage{color}
\usepackage{graphics}
\usepackage{geometry}
\usepackage{subcaption}
\usepackage{graphicx}
\usepackage{epstopdf}
\usepackage{dcolumn}
\usepackage[justification=centering]{caption}
\usepackage{natbib}   
\usepackage{setspace} 
\usepackage{multirow}
\usepackage{algorithm}
\usepackage{algorithmicx}
\usepackage{algpseudocode}
\usepackage{mathrsfs}
\usepackage{psfrag,epsf}
\usepackage{url} 
\hypersetup{colorlinks=true, linkcolor=red, urlcolor=blue, citecolor=blue}

\newtheorem{theorem}{Theorem}[section]
\newtheorem{proposition}[theorem]{Proposition}
\newtheorem{lemma}[theorem]{Lemma}
\newtheorem{claim}[theorem]{Claim}
\newtheorem{remark}{Remark}[section]
\def\hat{\widehat}
\newcommand{\argmin}{\mathop{\rm arg\min}}

\def\al{\boldsymbol{\alpha}}
\def\alp{\widetilde{\boldsymbol{\alpha}}}

\def\0{\boldsymbol{0}}

\def\G{\boldsymbol{G}}

\def\w{\boldsymbol{w}}

\def\x{\boldsymbol{x}}
\def\r{\boldsymbol{r}}

\def\R{\mathbb{R}}
\def\v{\boldsymbol{v}}
\def\V{\boldsymbol{V}}
\def\U{\textbf{U}}
\def\A{\boldsymbol{A}}
\def\C{\boldsymbol{C}}
\def\D{\boldsymbol{D}}

\def\z{\boldsymbol{z}}

\def\x{\boldsymbol{x}}

\def\be{\boldsymbol{\beta}}

\def\X{\boldsymbol{X}}

\def\r{\boldsymbol{r}}

\def\tee{\boldsymbol{\theta}}

\def\S{\boldsymbol{S}}

\def\0{\boldsymbol{0}}

\def\U{\boldsymbol{U}}

\def\x{\boldsymbol{x}}

\def\be{\boldsymbol{\beta}}

\def\X{\boldsymbol{X}}

\def\tX {\widetilde{\boldsymbol{X}}}
\def\tth {\widetilde{\boldsymbol{\theta}}}
\def\tb {\widetilde{\boldsymbol{\beta}}}
\def\ta {\widetilde{\boldsymbol{\alpha}}}
\def\eps {{\boldsymbol{\epsilon}}}

\definecolor{DSgray}{cmyk}{0,1,0,0}

\newcommand{\BlackBox}{\rule{1.5ex}{1.5ex}}

\begin{document}
	
	\title{Distributed Inference for Linear Support Vector Machine}
	
	\author{Xiaozhou Wang\footnote{Department of Mathematics, Institute of Natural Sciences and MOE-LSC, Shanghai Jiao Tong University, Email: wangxiaozhou@sjtu.edu.cn} ~ Zhuoyi Yang\footnote{Stern School of Business, New York University, Email: zyang@stern.nyu.edu.} ~  Xi Chen \footnote{Stern School of Business, New York University, Email: xchen3@stern.nyu.edu} ~ Weidong Liu \footnote{Department of Mathematics, Institute of Natural Sciences and MOE-LSC, Shanghai Jiao Tong University, Email: weidongl@sjtu.edu.cn}}
	
\date{}
	\maketitle
	\begin{abstract}
		
		The growing size of modern data brings many new challenges to existing statistical inference methodologies and theories, and calls for the development of distributed inferential approaches. This paper studies distributed inference for linear support vector machine (SVM) for the binary classification task. Despite a vast literature on SVM, much less is known about the inferential properties of SVM, especially in a distributed setting. In this paper, we propose a multi-round distributed linear-type (MDL) estimator for conducting inference for linear SVM. The proposed estimator is computationally efficient. In particular, it only requires an initial SVM estimator and then successively refines the estimator by solving simple weighted least squares problem. Theoretically, we establish the Bahadur representation of the estimator. Based on the representation, the asymptotic normality is further derived, which shows that the MDL estimator achieves the optimal statistical efficiency, i.e., the same efficiency as the classical linear SVM  applying to the entire data set in a single machine setup. Moreover, our asymptotic result avoids the condition on the number of machines or data batches, which is commonly assumed in distributed estimation literature, and allows the case of diverging dimension. We provide simulation studies to demonstrate the performance of the proposed MDL estimator.
	\end{abstract}
	

\section{Introduction}
\label{sec:intro}
The development of modern technology has enabled data collection of unprecedented size.
Very large-scale data sets, such as collections of images, text, transactional data, sensor network data, are becoming prevailing, with examples ranging from digitalized books and newspapers,  to collections of images on Instagram,  to data generated by large-scale networks of sensing devices or mobile robots. The scale of these data brings new challenges to traditional statistical estimation and inference methods, particularly in terms of memory restriction and computation time. For example, a large text corpus easily exceeds the memory limitation and thus cannot be loaded into memory all at once.
In a sensor network, the data are collected by each sensor in a distributed manner. It will incur an excessively high communication cost if we transfer all the data into a center for processing, and moreover, the center might not have enough memory to store all the data collected from different sensors.  In addition to memory constraints, these large-scale data sets also pose  challenges in computation. It will be computationally very expensive to directly apply an off-the-shelf optimization solver for computing the maximum likelihood estimator (or empirical risk minimizer) on the entire data set. These challenges call for new statistical inference approaches that are able to not only handle large-scale data sets efficiently, but also achieve the same statistical efficiency as classical approaches.

In this paper, we study the  problem of distributed inference for linear support vector machine (SVM). SVM, introduced by \cite{cortes1995support}, has been one of the most popular classifiers in statistical machine learning, which finds a wide range of applications in image analysis, medicine,  finance, and other domains. Due to the importance of SVM, various parallel SVM algorithms have been proposed in machine learning literature; see, e.g., \cite{graf2005parallel, forero2010consensus, zhu2008parallelizing, hsieh2014divide} and an overview in \cite{wang2012distributed}. However, these algorithms mainly focus on addressing the computational issue for SVM, i.e., developing a parallel optimization procedure to  minimize the objective function of SVM that is defined on given finite samples. In contrast, our paper aims to address the statistical inference problem, which is fundamentally different. More precisely, the task of distributed inference is to construct an estimator for the \emph{population risk minimizer} in a distributed setting and to characterize its \emph{asymptotic behavior} (e.g., establishing its limiting distribution).

As the size of data becomes increasingly large, distributed inference has received a lot of attentions and algorithms have been proposed for various problems (please see the related work Section \ref{sec:related} and references therein for more details). However, the problem of SVM possesses its own unique challenges in distributed inference. First, SVM is a classification problem that involves binary outputs $\{-1, 1\}$. Thus, as compared to regression problems, the noise structure in SVM is different and more complicated, which brings new technical challenges. We will elaborate this point with more details in Remark \ref{rmk:noise}. Second, the hinge loss in SVM is non-smooth. Third, instead of considering the fixed dimension $p$ as in many existing theories on asymptotic properties of SVM parameters \citep[see, e.g.,][]{lin1999some, zhang2004statistical, blanchard2008statistical, koo2008bahadur}, we aim to study the diverging $p$ case, i.e., $p \rightarrow \infty$ as the sample size $n \rightarrow \infty$.

\interfootnotelinepenalty=10000

To address aforementioned challenges, we focus ourselves on the distributed inference for linear SVM, as the first step to the study of distributed inference for more general SVM.\footnote{Our result relies on the Bahadur representation of the linear SVM estimator \citep[see, e.g.,][]{koo2008bahadur}. For general SVM, to the best of our knowledge, the Bahadur representation in a single machine setting is still open, which has to be developed before investigating distributed inference for general SVM. Thus, we leave this for future investigation.} Our goal is three-fold:
\begin{enumerate}
	\item The obtained estimator should achieve the same statistical efficiency as merging all the data together. That is, the distributed inference should not lose any statistical efficiency as compared to the ``oracle'' single machine setting.
	\item We aim to avoid any condition on the number of machines (or the number of data batches). Although this condition is widely assumed in distributed inference literature (see \citealp{lian2017divide} and Section \ref{sec:related} for more details), removing such a condition will make the results more useful in cases when the size of the entire data set is much larger than the memory size or in applications of sensor networks with a large number of sensors.
	\item The proposed algorithm should be computationally efficient. 	
\end{enumerate}

To simultaneously achieve these three goals, we develop a multi-round distributed linear-type (MDL) estimator for linear SVM. In particular,  by smoothing the hinge loss using a special kernel smoothing technique adopted from the quantile regression literature \citep{horowitz1998bootstrap,pang2012variance,chen2018quantile}, we first introduce a linear-type estimator  in a single machine setup. Our linear-type estimator requires a consistent initial SVM estimator that can be easily obtained by solving SVM on one local machine. Given the initial estimator $\tb_0$, the linear-type estimator has a simple and explicit formula that greatly facilitates the distributed computing. Roughly speaking, given $n$ samples $(y_i, \X_i)$ for $i=1,\ldots, n$, our linear-type estimator takes the form of ``weighted least squares'':
\begin{equation}\label{eq:linear}
\tb =\Bigl[\underbrace{\frac{1}{n}\sum_{i=1}^n u_i(y_i, \X_i, \tb_0) \X _i\X _i^\mathrm{T}}_{A_1}\Bigr]^{-1} \Bigl\{\underbrace{\frac{1}{n}\sum_{i=1}^n v_i(y_i, \X_i, \tb_0)y_i\X _i - \w(\tb_0)}_{A_2}\Bigr\},
\end{equation}
where the term $A_1$ is a weighted gram matrix and $u_i(y_i, \X_i, \tb_0) \in \mathbb{R}$ is the weight that only depends on the $i$-th data $(y_i, \X_i)$ and $\tb_0$. In the vector $A_2$, $\w(\tb_0)$ is a fixed vector that only depends on $\tb_0$ and $v_i(y_i, \X_i, \tb_0) \in \mathbb{R}$ is the weight that only depends on $(y_i, \X_i, \tb_0)$. The formula in \eqref{eq:linear} has a similar structure as weighted least squares, and thus can be easily computed in a distributed environment (noting that each term in $A_1$ and $A_2$ only involves the $i$-th data point $(y_i, \X_i)$ and there is no interaction term in Equation \ref{eq:linear}). In addition, the linear-type estimator in \eqref{eq:linear} can be efficiently computed by solving a linear equation system (instead of computing matrix inversion explicitly), which is computationally more attractive than solving the non-smooth optimization in the original linear SVM formulation.

The linear-type estimator can easily refine itself by using the $\tb$ on the left hand side of \eqref{eq:linear} as the initial estimator. In other words, we can obtain a new linear-type estimator by recomputing the right hand side of \eqref{eq:linear} using $\tb$ as the initial estimator. By successively refining the initial estimator for $q$ rounds/iterations, we could obtain the final  multi-round distributed linear-type (MDL) estimator  $\tb^{(q)}$.  The estimator $\tb^{(q)}$ not only has its advantage in terms of computation in a distributed environment, but also has describable statistical properties. In particular, with a small number $q$, the estimator $\tb^{(q)}$ is able to achieve the optimal statistical efficiency, that is, the same efficiency as the classical linear SVM estimator computed on the entire data set. To establish the limiting distribution and statistical efficiency results, we first develop the Bahadur representation of our MDL estimator  of SVM (see Theorem \ref{thm:bahadur}). Then the asymptotic normality follows immediately from the Bahadur representation.
It is worthwhile noting that the Bahadur representation \citep[see, e.g.,][]{bahadur1966note,koenker1978regression,chaudhuri1991nonparametric} provides an important characterization of the asymptotic behavior of an estimator.  For the original linear SVM formulation, \cite{koo2008bahadur} first established the Bahadur representation. In this paper, we establish the Bahadur representation of our multi-round distributed linear-type estimator.

Finally, it is worthwhile noting  that our algorithm is similar to a recently developed algorithm for distributed quantile regression \citep{chen2018quantile}, where both algorithms rely on a kernel smoothing technique and linear-type estimators. However, the technique for establishing the theoretical property for linear SVM is quite different from that for quantile regression. The difference and new technical challenges in linear SVM will be illustrated in Remark \ref{rmk:noise} (see Section \ref{sec:method}).

The rest of the paper is organized as follows. In Section \ref{sec:related}, we provide a brief overview of related works. Section \ref{sec:method} first introduces the problem setup and then describes the proposed linear-type estimator and MDL estimator for linear SVM. In Section \ref{sec:theory}, the main theoretical results are given. Section \ref{sec:sim} provides the simulation studies to illustrate the performance of MDL estimator of SVM. Conclusions and future works are given in Section \ref{sec:conclusion}. We provide the proofs of our theoretical results in Appendix \ref{app:theorem}.

\section{Related Works}
\label{sec:related}

In distributed inference literature, the divide-and-conquer (DC) approach is one of the most popular approaches and has been applied to a wide range of statistical problems. In the standard DC framework, the entire data set of $n$ i.i.d. samples is evenly split into $N$ batches or distributed on $N$ local machines. Each machine computes a local estimator using the $m=n/N$ local samples. Then, the final estimator is obtained by averaging local estimators. The performance of the DC approach  (or its variants)  has been investigated on many statistical problems, such as density parameter estimation \citep{li2013statistical},  kernel ridge regression \citep{zhang2015divide},  high-dimensional linear regression \citep{lee2017communication} and generalized linear models \citep{chen2014split, battey2015distributed}, semi-parametric partial linear models \citep{zhao2016partially}, quantile regression \citep{volgushev2017distributed, chen2018quantile},  principal component analysis \citep{Fan17distributed},  one-step estimator \citep{HuangHuo:15}, high-dimensional SVM \citep{lian2017divide}, $M$-estimators with cubic rate \citep{shi2016massive}, and some non-standard problems where rates of convergence are slower than $n^{1/2}$ and limit distributions are non-Gaussian \citep{banerjee2016divide}. On one hand, the DC approach enjoys low communication cost since it only requires one-shot communication (i.e., taking the average of local estimators). On the other hand, almost all the existing work on DC approaches requires a constraint on the number of machines. The main reason is that the averaging only reduces the variance but not the bias of each local estimator. To make the variance the dominating term in the final estimator constructed by taking averaging, the constraint on the number of machines is unavoidable. In particular, in the DC approach for linear SVM in \cite{lian2017divide}, the number of machines $N$ has to satisfy the condition $N\leq (n/\log(p))^{1/3}$ \citep[see Remark 1 in][]{lian2017divide}. As a comparison, our MDL estimator that involves multi-round aggregations successfully eliminates this condition on the number of machines.

In fact, to relax this constraint, several multi-round distributed methods have been recently developed \cite[see][]{wang2016efficient,jordan2016communication}.
In particular, the key idea behind these methods is to approximate the Newton step by using the local Hessian matrix computed on a local machine.  However, to compute the local Hessian matrix, their methods require the second-order differentiability on the loss function and thus are not applicable to problems involving non-smooth loss such as SVM.

The second line of the related research is the support vector machine (SVM). Since it was proposed by \cite{cortes1995support}, there is a large body of literature on SVM from both machine learning and statistics community. The readers might refer to the books \citep{cristianini2000introduction,scholkopf2002learning,steinwart2008support} for a comprehensive review of SVM. In this section, we briefly mention a few relevant works on the statistical properties of linear SVM. In particular, the Bayes risk consistency and the rate of convergence of SVM have been extensively investigated \citep[see, e.g.,][]{lin1999some,zhang2004statistical,blanchard2008statistical,bartlett2006convexity}. These works mainly concern the asymptotic risk. For the asymptotic properties of underlying coefficients, \cite{koo2008bahadur} first established the Bahadur representation of linear SVM under the fixed $p$ setting. \cite{jiang2008estimating} proposed interval estimators for the prediction error for general SVM. For the large $p$ case, there are two common settings. One assumes that $p$ grows to infinity at a slower rate than (or linear in) the sample size $n$ but without any sparsity assumption. Our paper also belongs to this setup. Under this setup, \cite{huang2017asymptotic} investigated the angle between the normal direction vectors of SVM separating hyperplane and corresponding Bayes optimal separating hyperplane under spiked population models. Another line of research considers high-dimensional SVM under a certain sparsity assumption on underlying coefficients. Under this setup, \cite{peng2016error} established the error bound in $L_1$ norm. \cite{zhang2016consistent} and \cite{zhang2016variable} investigated the variable selection problem in linear SVM.

\section{Methodology}
\label{sec:method}
\subsection{Preliminaries}
\label{sec:pre}

In a standard binary classification problem setting, we consider a pair of random variables $\{\X,Y\}$ with $\X \in \mathcal{X} \subseteq \mathbb{R}^p$ and $Y \in \{-1,1\}$. The marginal distribution of $Y$ is given by $\mathbb{P}(Y=1)=\pi_+$ and $\mathbb{P}(Y=-1)=\pi_-$ where $\pi_+,\pi_- >0 $ and $\pi_++\pi_- =1$. We assume that the random vector $\X$ has a continuous distribution on $\mathcal{X}$ given $Y$. Let $\{\X_i,y_i\}_{i=1,...,n}$ be i.i.d. samples drawn from the joint distribution of random variables $\{\X,Y\}$. In the linear classification problem, a hyperplane is defined by $\beta_0+\X^\mathrm{T}\be = 0$ with $\be = (\beta_1,\beta_2,...,\beta_p)^\mathrm{T}$. Define $\tX = (1,X_1,...,X_p)^\mathrm{T}$ and the coefficient vector $\tb = (\beta_0,\beta_1,...,\beta_p)^\mathrm{T}$. For convenience purpose we also define $l(\X;\tb )=\beta_0+\X^\mathrm{T}\be=\tX ^\mathrm{T}
\tb $. In this paper we consider the standard non-separable SVM formulation, which takes the following form
\begin{equation}\label{opt}
f_{\lambda,n} (\tb) = \frac{1}{n} \sum_{i=1}^n\left(1-y_il(\X_i;\tb )\right)_++\frac{\lambda}{2}\|\be\|^{2}_{2},
\end{equation}
\begin{equation}\label{eqn:all}
\tb _\text{SVM all} = \mathop{\arg\min}_{\tb \in\mathbb{R}^{p+1}} f_{\lambda,n} (\tb).
\end{equation}
Here $(u)_+ = \max(u,0)$ is the hinge loss, $\lambda>0$ is the regularization parameter and $\|\cdot\|_{2}$ denotes the Euclidean norm of a vector. We note that we do not penalize the first coordinate $\beta_0$ and thus the regularization is only imposed on $\be$ instead of $\widetilde{\be}$. Throughout this paper, for any $(p+1)$-dimensional parameter vector $\widetilde{\al}$, we will use $\al$ to denote the subvector of  $\widetilde{\al}$ without the first coordinate, and only $\al$ will appear in the regularization term.

The corresponding population loss function is defined as
\[
L(\tb) = \mathbb{E}[1-Yl(\X;\tb )]_+.
\]

We denote the minimizer for the population loss by
\begin{equation}\label{eqn:pop}
\tb ^*=\mathop{\arg\min}_{\tb \in\mathbb{R}^{p+1}}\mathbb{E}[1-Yl(\X;\tb )]_+.
\end{equation}
\cite{koo2008bahadur} proved that under some mild conditions (see \citealp{koo2008bahadur} Theorem 1,2), there exists a unique minimizer for \eqref{eqn:pop} and it is nonzero (i.e., $\tb^* \neq \boldsymbol{0}$). We assume that these conditions hold throughout the paper. The  minimizer $\tb^*$ of the population loss function  will serve as the ``true parameter'' in our estimation problem and the goal is to construct an estimator and make inference of $\tb^*$.
We further define some useful quantities as follows:
\[
\epsilon = 1-Yl(\X,\tb^*),\text{ and }\epsilon_i = 1-y_il(\X_i,\tb^*).
\]
The reason why we use the notation $\epsilon$ is because it plays a similar role in the theoretical analysis as the noise term in a standard regression problem. However, as we will show in Section \ref{sec:method} and \ref{sec:theory}, the behavior of $\epsilon$ is quite different from the noise in a classical regression setting since it does not have a continuous density function (see Remark \ref{rmk:noise}). Next, denote by $\delta(\cdot)$ the Dirac delta function, we define
\begin{equation}\label{eqn:SD}
\begin{aligned}
\S(\tb )&=-\mathbb{E}[I{\{1-Y\tX ^\mathrm{T}\tb \geq0\}}Y\tX ],\\
\D(\tb )&=\mathbb{E}[\delta(1-Y\tX ^\mathrm{T}\tb )\tX \tX ^\mathrm{T}],
\end{aligned}
\end{equation}
where $I\{\cdot\}$ is the indicator function.

The quantities $\S(\tb )$ and $\D(\tb )$ can be viewed as the gradient and Hessian matrix of $L(\tb)$ and we assume that the smallest eigenvalue of $\D(\tb^* )$ is bounded away from 0. In fact these assumptions can be verified under some regular conditions (see \citealp{koo2008bahadur} Lemma 2, Lemma 3 and Lemma 5 for details)
and are common in SVM literature (e.g., \citealp{zhang2016variable} Condition 2 and 6).
\subsection{A Linear-type Estimator for SVM}
\begin{figure}[t!]
	\centering
	\includegraphics[width=0.6\textwidth]{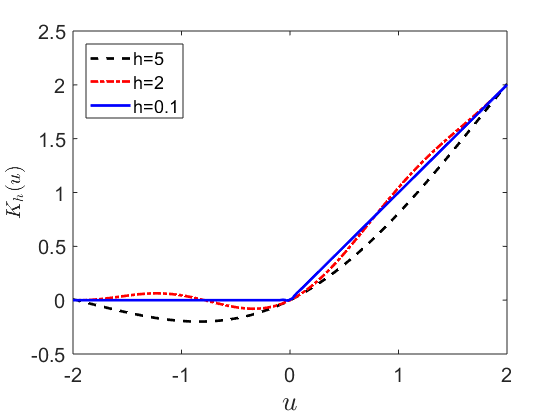}
	\caption{An example of the smoothed hinge loss function $K_h$ with different bandwidth $h$.\\ See Section \ref{sec:sim} for details in the construction of $H(\cdot)$.} 
	\label{fig:loss}
\end{figure}

In this section, we first propose a linear-type estimator for SVM on a single machine which can be later extended to a distributed algorithm. The main challenge in solving the optimization problem in \eqref{opt} is that the objective function is non-differentiable due to the appearance of hinge loss. Motivated by a smoothing technique from quantile regression literature (see, e.g., \citealp{chen2018quantile,horowitz1998bootstrap,pang2012variance}), we consider a smooth function $H(\cdot)$ satisfying $H(u)=1$ if $u\geq1$ and $H(u) = 0$ if $u\leq -1$. We replace the hinge loss with its smooth approximation $K_h(u) = uH(\frac{u}{h})$, where $h$ is the bandwidth. As the bandwidth $h\to 0$, $H(\frac{u}{h})$ and $\frac{1}{h}H'(\frac{u}{h})$ approaches the indicator function $I\{u\geq 0\}$ and Dirac delta function $\delta(u)$ respectively, and $K_h(u)$ approximates the hinge loss $\max(u,0)$ (see Figure \ref{fig:loss} for an example of $K_h$ with different bandwidths). To motivate our linear-type estimator, we first consider the following estimator with the non-smooth hinge loss in linear SVM replaced by its smooth approximation:
\begin{equation}\label{eqn:smooth}
\begin{aligned}
\tb _h&=\mathop{\arg\min}_{\tb \in\mathbb{R}^{p+1}}\frac{1}{n}\sum_{i=1}^n\left[1-y_il(\X_i;\tb )\right]H\left(\frac{1-y_il(\X_i;\tb )}{h}\right)+\frac{\lambda}{2}\lVert\be\rVert^2_{2}\\
&=\mathop{\arg\min}_{\tb \in\mathbb{R}^{p+1}}\frac{1}{n}\sum_{i=1}^n K_h(1-y_il(\X_i;\tb ))+\frac{\lambda}{2}\lVert\be\rVert^2_{2}.
\end{aligned}
\end{equation}
Since the objective function is differentiable and $\frac{dK_h(x)}{dx}=H(x/h)+\frac{x}{h}H'(x/h)$, by the first order condition (i.e., setting the derivative of the objective function in \eqref{eqn:smooth} to zero), $\tb_h$ satisfies
\begin{equation*}
\frac{1}{n}\sum_{i=1}^n(-y_i\tX _i)\left[H\left(\frac{1-y_il(\X_i;\tb _h)}{h}\right)+\frac{1-y_il(\X_i;\tb _h)}{h}H'\left(\frac{1-y_il(\X_i;\tb _h)}{h}\right)\right]+\lambda \binom{0}{\be_h}=0.
\end{equation*}
We first rearrange the equation and express $\tb_h$ by
\begin{equation}\label{eq:fixed}
\begin{aligned}
\tb _h=&\left[\frac{1}{n}\sum_{i=1}^n \tX _i\tX _i^\mathrm{T}\frac{1}{h}H'\left(\frac{1-y_il(\X_i;\tb _h)}{h}\right)\right]^{-1}\\&\times \left\{\frac{1}{n}\sum_{i=1}^n y_i\tX _i\left[H\left(\frac{1-y_il(\X_i;\tb _h)}{h}\right)+\frac{1}{h}H'\left(\frac{1-y_il(\X_i;\tb _h)}{h}\right)\right]-\lambda \binom{0}{\be_h}\right\}.
\end{aligned}
\end{equation}
This fixed-point form formula for $\tb_h$ cannot be solved explicitly since $\tb_h$ appears on both sides of \eqref{eq:fixed}. Nevertheless, $\tb_h$ is not our final estimator and is mainly introduced to motivate our estimator. The key idea is to replace $\tb _h$ on the right hand side of \eqref{eq:fixed} by a consistent initial estimator $\tb _0$ (e.g., $\tb_0 $ can be constructed by solving a linear SVM on a small batch of samples). Then, we obtain the following linear-type estimator for $\tb ^*$:
\begin{equation}\label{eqn:LE}
\begin{aligned}
\tb =&\left[\frac{1}{n}\sum_{i=1}^n \tX _i\tX _i^\mathrm{T}\frac{1}{h}H'\left(\frac{1-y_il(\X_i;\tb _0)}{h}\right)\right]^{-1}\\&\times \left\{\frac{1}{n}\sum_{i=1}^n y_i\tX _i\left[H\left(\frac{1-y_il(\X_i;\tb _0)}{h}\right)+\frac{1}{h}H'\left(\frac{1-y_il(\X_i;\tb _0)}{h}\right)\right]-\lambda \binom{0}{\be_{0}}\right\}.
\end{aligned}
\end{equation}
Notice that \eqref{eqn:LE} has a similar structure as weighted least squares (see the explanations in the paragraph below \eqref{eq:linear} in the introduction). As shown in the following section, this weighted least squares formulation can be computed efficiently in a distributed setting.
\subsection{Multi-Round Distributed Linear-type (MDL) Estimator}
It is important to notice that given the initial estimator $\tb_0$, the linear-type estimator in \eqref{eqn:LE} only involves summation of matrices and vectors computed for each individual data point. Therefore based on \eqref{eqn:LE}, we will construct a multi-round distributed linear-type estimator (MDL estimator) that can be efficiently implemented in a distributed setting.

First, let us assume that the total data indices $\{1,...,n\}$ are divided into $N$ subsets $\{\mathcal{H}_1,...,\mathcal{H}_N\}$ with equal size $m = n/N$. Denote by $\mathcal{D}_k = \{(\X_i,y_i):i\in \mathcal{H}_k \}$ the data in the $k$-th local machine. In order to compute $\tb$, for each batch of data $\mathcal{D}_k$ for $k=1,...,N$, we define the following quantities

\begin{equation}\label{eqn:UV}
\begin{aligned}
\U_k&=\frac{1}{n}\sum_{i\in\mathcal{H}_k} y_i\tX _i\left[H\left(\frac{1-y_il(\X_i;\tb _0)}{h}\right)+\frac{1}{h}H'\left(\frac{1-y_il(\X_i;\tb _0)}{h}\right)\right],\\
\V_k&=\frac{1}{n}\sum_{i\in\mathcal{H}_k}\tX _i\tX _i^\mathrm{T}\frac{1}{h}H'\left(\frac{1-y_il(\X_i;\tb _0)}{h}\right).
\end{aligned}
\end{equation}

Given $\tb_0$, the quantities $\U_k,\V_k$ can be computed independently in each machine and only $(\U_k,\V_k)$ has to be stored and transferred to the central machine. Then after receiving $(\U_k,\V_k)$ from all the machines, the central machine can aggregate the data and compute the estimator by
\[
\tb^{(1)} = \left(\sum_{k=1}^N \V_k \right)^{-1}\left(\sum_{k=1}^N \U_k-\lambda \binom{0}{\be_0}\right).
\]
Then $\tb^{(1)}$ can be sent to all the machines to repeat the whole process to construct $\tb^{(2)}$ using $\tb^{(1)}$ as the new initial estimator. The algorithm is repeated $q$ times for a pre-specified $q$ (see Equation \ref{eqn:q} for details), and $\tb^{(q)}$ is taken to be the final estimator (see Algorithm \ref{algo:dsvm} for details). We name this estimator as the multi-round distributed linear-type (MDL) estimator.

We notice that instead of computing matrix inversion $ \left(\sum_{k=1}^N \V_k \right)^{-1}$ in every iteration which has a computation cost $O(p^3)$, one only needs to solve a linear system in \eqref{eq:al}. Linear system has been studied in numeric optimization for several decades and many efficient algorithms have been developed, such as conjugate gradient method \citep{hestenes1952methods}. We also notice that we only have to solve a single optimization problem on one local machine to compute the initial estimator. Then at each iteration, only matrix multiplication and summation needs to be computed locally which makes the algorithm computationally efficient. It is worthwhile noticing that according to Theorem \ref{thm:asym} in Section \ref{sec:theory}, under some mild conditions, if we choose $h:=h_g=\max\left(\lambda,\sqrt{p/n},(p/m)^{2^{g-2}}\right)$ for $1\le g\le q$, the MDL estimator $\tb^{(q)}$ achieves optimal statistical efficiency as long as $q$ satisfies \eqref{eqn:q}, which is usually a small number. Therefore, a few rounds of iterations would guarantee good performance for the MDL estimator.

\begin{algorithm}[!t]
	\caption{Multi-round distributed linear-type estimator (MDL) for SVM}
	\label{algo:dsvm}
	\begin{algorithmic}[1]
		\Require
		Samples stored in the machines $\{\mathcal{D}_1,...,\mathcal{D}_N\}$, the number of iterations $q$, smooth function $H$, bandwidths $\{h_1,...,h_q\}$ and regularization parameter $\lambda$.
		\For{$g=1,\ldots,q$}
		\If{$g=1$}
		\State Compute the initial estimator based on $\mathcal{D}_1$:
		\begin{equation*}
		\tb_0 = \mathop{\arg\min}_{\tb \in\mathbb{R}^{p+1}}\frac{1}{m} \sum_{i\in\mathcal{H}_1}\left(1-y_il(\X_i;\tb )\right)_+.
		\end{equation*}
		\Else
		\State $\tb _0=\tb ^{(g-1)}$
		\EndIf
		\State $\tb_0$ is transferred to all the local machines.
		\For{$k=1,\ldots,N$}
		\State Compute $(\U_k,\V_k)$ according to \eqref{eqn:UV} with data in $\mathcal{D}_k$ using the bandwidth $h_g$.
		\State Transfer $(\U_k,\V_k)$ to the central machine.
		\EndFor
		\State The central machine computes the estimator $\tb ^{(g)}$ by
		\begin{equation}\label{eq:al}
		\tb ^{(g)}=\left(\sum_{k=1}^N \V_k\right)^{-1}\left(\sum_{k=1}^N \U_k-\lambda \binom{0}{\be_0}\right).
		\end{equation}
		\EndFor
		\Ensure
		The final MDL estimator $\tb ^{(q)}$.
	\end{algorithmic}
\end{algorithm}

For the choice of the initial estimator in the first iteration, we propose to construct it by solving the original SVM optimization \eqref{opt} only on a small batch of samples (e.g., the samples on the first machine $\mathcal{D}_1$). The estimator $\tb_0$ is only a crude estimator for $\tb^*$, but we will prove later that it is enough for the algorithm to produce an estimator with optimal statistical efficiency under some regularity conditions. In particular, if we compute the initial estimator in the first round on the first batch of data, we will solve the following optimization problem
\[
\tb_0 = \mathop{\arg\min}_{\tb \in\mathbb{R}^{p+1}}\frac{1}{m} \sum_{i\in\mathcal{H}_1}\left(1-y_il(\X_i;\tb )\right)_+.
\]
Then we have the following proposition from \cite{zhang2016variable}.
\begin{proposition}[\cite{zhang2016variable}]\label{prop:init}
	Under conditions (C1)-(C6) in \cite{zhang2016variable}, we have
	\[
	\|\tb_0-\tb^*\|_2 = O_\mathbb{P}(\sqrt{p/m}).
	\]
\end{proposition}
According to our Theorem \ref{thm:bahadur}, the initial estimator $\tb_0$ needs to satisfy $\|\tb_0-\tb^*\|_2 = O_\mathbb{P}(\sqrt{p/m})$, and therefore the estimator computed on the first machine is a valid initial estimator. On the other hand, one can always use different approaches to construct the initial estimator $\tb_0$ as long as it is a consistent estimator.

\begin{remark}\label{rmk:noise}
	\normalfont
	We note that although Algorithm \ref{algo:dsvm} has a similar form as the DC-LEQR estimator for quantile regression (QR) in \cite{chen2018quantile}, the structures of the SVM and QR problems are fundamentally different and thus the theoretical development for establishing the Bahadur representations for SVM is more challenging. To see that, let us recall the quantile regression model:
	\begin{equation}\label{eq:QR}
	Y=\tX^\mathrm{T} \tb^* +\epsilon,
	\end{equation}
	where $\epsilon$ is the unobserved random noise satisfying $\Pr(\epsilon \leq 0 |\tX)=\tau$ and $\tau$ is known as the quantile level. The asymptotic results of QR estimators heavily rely on the Lipschitz continuity assumption on the conditional density $f(\epsilon|\tX)$ of $\epsilon$ given $\tX$, which has been assumed in almost all existing literature. In the SVM problem, the quantity $\epsilon:=1-Y\tX^\mathrm{T} \tb^*$ plays a similar role as the noise in a regression problem. However, since $Y \in \{-1,1\}$ is binary, the conditional distribution $f(\epsilon|\tX)$ becomes a two-point distribution, which no longer has a density function. To address this challenge and derive the asymptotic behavior of SVM, we directly work on the joint distribution of $\epsilon$ and $\tX$. As the dimension of  $\tX$ (i.e., $p+1$) can go to infinity, we use a slicing technique by considering the one-dimensional marginal distribution of $\tX$ (see Condition (C2) and proof of Theorem \ref{thm:bahadur} for more details).
\end{remark}

\subsection{Communication-Efficient Implementation}\label{sec:efficient}
In this section, we discuss a communication-efficient implementation of the proposed MDL estimator. Note that in Algorithm \ref{algo:dsvm}, each local machine transmits a $(p+1)$-by-$(p+1)$ matrix $\V_k$ to the central machine at each iteration. In fact, the communication of $(p+1)$-by-$(p+1)$ matrices can be avoided by using the approximate Newton method (see, e.g., \citealp{shamir2014communication,wang2016efficient,jordan2016communication}). Instead of transmitting the Hessian matrix $\V_k$, we will only use the local Hessian matrix $\V_1$ computed on the first machine. More specifically, the estimator $\tb^{(g)}$ in \eqref{eq:al} essentially solves the minimization problem
\begin{equation}\label{eqn:newtonloss}
\begin{aligned}
\argmin_{\tb\in \R^{p+1}}\frac{1}{2}\tb^\mathrm{T} \left(\sum_{k=1}^N \V_k\right)\tb -\tb^\mathrm{T}\left(\sum_{k=1}^N\U_k-\lambda \binom{0}{\be_0}\right).
\end{aligned}
\end{equation}
The approximate Newton method uses the following iterations to solve the above minimization problem:
\begin{equation}\label{eq:surrogate_onestep}
\begin{aligned}
\tb^{(g,t)} = \tb^{(g,t-1)}-\left(N\hat{\V}_1\right)^{-1}\left(\sum_{k=1}^N\left(\V_k\tb^{(g,t-1)}-\U_k\right)+\lambda \binom{0}{\be_0}\right),\quad \tb^{(g,0)} = \tb_{0},
\end{aligned}
\end{equation}	
as an inner iterative procedure, which approximately solves the equation \eqref{eq:al}.  In \eqref{eq:surrogate_onestep}, we let $\hat{\V}_{1}=\V_{1}$  in \eqref{eqn:UV} with the bandwidth $h=\sqrt{p/m}$.  The matrix $N\hat{\V}_{1}$ is used to approximate the Hessian matrix $\sum_{k=1}^{N}\V_{k}$ in \eqref{eqn:newtonloss}.
To compute the minimizer of \eqref{eq:surrogate_onestep},  we note that the matrix $\hat{\V}_1$ only involves the data on the first machine, and thus there is no need to communicate $(p+1)$-by-$(p+1)$ matrices to compute \eqref{eq:surrogate_onestep}. In fact, each local machine only transmits a $(p+1)$-by-1 vector $\V_k\tb^{(g,t-1)}-\U_k$ to the central machine.
We present the entire communication-efficient implementation in Algorithm \ref{algo:dsvm_efficient}.

Recall that $\tb^{(g)}$ is the estimator defined in \eqref{eq:al}. It is easy to show that  $\tb^{(g,t)}$ in \eqref{eq:surrogate_onestep} converges  to $\tb^{(g)}$ at a super-linear rate:
\[
\|	\tb^{(g,t)}-\tb^{(g)}\|_2 \leq\Big{\|}I-\hat{\V}^{-1}_{1}\frac{1}{N}\sum_{k=1}^{N}\V_{k}\Big{\|}	\|	\tb^{(g,t-1)}-\tb^{(g)}\|_2,
\]
where $\|\cdot\|$ denotes the spectral norm of a matrix.
By repeatedly applying the argument, we have the following proposition, whose proof is relegated to Appendix \ref{app:aux}.
\begin{proposition}
	\label{prop:efficient}
	Assume that the conditions of Theorem \ref{thm:bahadur} in Section \ref{sec:theory} hold.  Suppose that $p=O(m^{\nu})$ for some $0<\nu<1$ and $n=O(m^{A})$ for some $A>0$. We have
	\begin{equation}\label{eqn:efficient}
	\|\tb^{(g,t)} - \tb^{(g)}\|_2  = O_\mathbb{P}(m^{-\delta t})
	\end{equation}
	for some constant $\delta>0$, where $\delta$ and $O_\mathbb{P}(1)$ do not depend on $t$.
\end{proposition}
Proposition \ref{prop:efficient} and the convergence rate of $\tb^{(g)}$ (see Theorem \ref{thm:bahadur}) imply that the inner procedure takes at most constant-valued iterations to achieve the same convergence rate as $\tb^{(g)}$. Therefore the theoretical results of the MDL estimator in Algorithm \ref{algo:dsvm} (see Theorem \ref{thm:bahadur} and \ref{thm:asym} in Section \ref{sec:theory}) still hold for Algorithm \ref{algo:dsvm_efficient}. In summary, as compared to Algorithm \ref{algo:dsvm}, Algorithm \ref{algo:dsvm_efficient} only requires $O(p)$ communication cost  for each local machine. Therefore, Algorithm \ref{algo:dsvm_efficient} is communicationally more efficient when $p$ is large.
\begin{algorithm}[!t]
	\caption{Communication-efficient MDL for SVM}
	\label{algo:dsvm_efficient}
	\begin{algorithmic}[1]
		
		\Require
		Samples stored in the machines $\{\mathcal{D}_1,...,\mathcal{D}_N\}$, the number of outer iterations $q$, the number of inner iterations $T$, smooth function $H$, bandwidths $\{h_1,...,h_q\}$ and regularization parameter $\lambda$.
		\For{$g=1,\ldots,q$}
		\If{$g=1$}
		\State Compute the initial estimator based on $\mathcal{D}_1$:
		\begin{equation*}
		\tb_0 = \mathop{\arg\min}_{\tb \in\mathbb{R}^{p+1}}\frac{1}{m} \sum_{i\in\mathcal{H}_1}\left(1-y_il(\X_i;\tb )\right)_+.
		\end{equation*}
		\Else
		\State $\tb_0=\tb ^{(g-1)}$
		\EndIf
		\State Let $\tb^{(g,0)} = \tb_0$.
		\For{$t = 1,\ldots,T$}
		\State $\tb^{(g,t-1)}$ is transferred to all the local machines.
		\For{$k=1,\ldots,N$}
		\State Compute $\V_k\tb^{(g,t-1)} -\U_k$ and transfer it to the central machine.
		\EndFor
		\State The central machine computes $\tb ^{(g,t)}$ by
		\[
		\tb^{(g,t)} = \tb^{(g,t-1)}-\left(N\hat{\V}_1\right)^{-1}\left(\sum_{k=1}^N\left(\V_k\tb^{(g,t-1)}-\U_k\right)+\lambda \binom{0}{\be_0}\right),
		\]
		\State where $\hat{\V}_1$ is defined as $\V_1$ in \eqref{eqn:UV} but with the bandwidth $h = \sqrt{p/m}$.
		\EndFor
		\State Let $\tb^{(g)} = \tb^{(g,T)}$.
		\EndFor
		\Ensure
		The final MDL estimator $\tb ^{(q)}$.
		
	\end{algorithmic}
\end{algorithm}

\section{Theoretical Results}\label{sec:theory}
In this section, we give a Bahadur representation of the MDL estimator $\tb ^{(q)}$ and establish its asymptotic normality result. From \eqref{eqn:LE}, the difference between the MDL estimator and the true coefficient can be written as
\begin{equation}\label{eqn:beta}
\tb -\tb ^*=\D_{n,h}^{-1}\left(\A_{n,h}-\lambda \binom{0}{\be_{0}}\right),
\end{equation}
where $\A_{n,h}=\A_{n,h}(\tb _0)$, $\D_{n,h}=\D_{n,h}(\tb _0)$, and for any $\ta$, $\A_{n,h}(\ta)$ and $\D_{n,h}(\ta)$ are defined as follows,

\begin{equation*}
\begin{aligned}
\A_{n,h}(\ta )&=\frac{1}{n}\sum_{i=1}^n y_i\tX _i\left[H\left(\frac{1-y_il(\X _i;\ta) }{h}\right)+\frac{1-y_il(\X _i;\tb^*)}{h}H'\left(\frac{1-y_il(\X _i;\ta)}{h}\right)\right],\\
\D_{n,h}(\ta )&=\frac{1}{nh}\sum_{i=1}^n\tX _i\tX _i^\mathrm{T}H'\left(\frac{1-y_il(\X _i;\ta) }{h}\right).
\end{aligned}
\end{equation*}

For a good initial estimator $\tb_0$ which is close to $\tb^*$, the quantities $\A_{n,h}(\tb_0)$ and $\D_{n,h}(\tb_0)$ are close to $\A_{n,h}(\tb^*)$ and $\D_{n,h}(\tb^*)$. Recall that $\epsilon_i = 1-y_i\tX_i^\mathrm{T}\tb^*$ and we have
\begin{equation*}
\begin{aligned}
\A_{n,h}(\tb^* )&=\frac{1}{n}\sum_{i=1}^n y_i\tX _i\left[H\left(\frac{\epsilon_i }{h}\right)+\frac{\epsilon_i}{h}H'\left(\frac{\epsilon_i}{h}\right)\right],\\
\D_{n,h}(\tb^* )&=\frac{1}{nh}\sum_{i=1}^n\tX _i\tX _i^\mathrm{T}H'\left(\frac{\epsilon_i }{h}\right).
\end{aligned}
\end{equation*}
When $h$ is close to zero, the term $H\left(\frac{\epsilon_i }{h}\right)+\frac{\epsilon_i}{h}H'\left(\frac{\epsilon_i}{h}\right)$ in parenthesis of $\A_{n,h}(\tb^*)$ approximates $I\{\epsilon_i \geq 0  \}$. Therefore, $\A_{n,h}(\tb^* )$ will be close to $\frac{1}{n}\sum_{i=1}^n y_i\tX _i I\{\epsilon_i \geq 0  \}$. Moreover, since $\frac{1}{h}H'(\cdot/h)$ approximates Dirac delta function as $h \rightarrow 0$, $\D_{n,h}(\tb^*)$ approaches $$\D_n(\tb^*)= \frac{1}{n}\sum_{i=1}^n[\delta(\epsilon_i)\tX_i\tX_i^\mathrm{T}].$$
When $n$ is large, $\D_n(\tb^*)$ will be close to its corresponding population quantity $\D(\tb^*) = \mathbb{E}[\delta(\epsilon )\tX \tX ^\mathrm{T}]$ defined in \eqref{eqn:SD}.

According to the above argument, when $\tb_0$ is close to $\tb^*$, $\A_{n,h}(\tb_0)$ and $\D_{n,h}(\tb_0)$ approximate $\frac{1}{n}\sum_{i=1}^n y_i\tX _i I\{\epsilon_i \geq 0  \}$ and $\D(\tb^*)$, respectively. Therefore, by \eqref{eqn:beta},  we would expect $\tb-\tb^*$ to be close to the following quantity,
\begin{equation}\label{eqn:bahamain}
\begin{aligned}
\D(\tb ^*)^{-1}\left(\frac{1}{n}\sum_{i=1}^ny_i\tX _iI\{\epsilon_i\ge0\}-\lambda \binom{0}{\be^{*}}\right).
\end{aligned}
\end{equation}

We will see later that \eqref{eqn:bahamain} is exactly the main term of the Bahadur representation of the estimator. Next, we formalize these statements and present the asymptotic properties of $\A_{n,h}$ and $\D_{n,h}$ in Proposition \ref{prop:C} and \ref{prop:D}. The asymptotic properties of the MDL estimator will be provided in Theorem \ref{thm:bahadur}. To this end, we first introduce some notations and assumptions for the theoretical result.

Recall that $\be^*=(\beta_1^*,...,\beta_p^*)^\mathrm{T}$ and for $\X=(X_1,...,X_p)^\mathrm{T}$, let $\X_{-s}$ be a ($p-1$)-dimensional vector with $X_s$ removed from $\X$. Similar notations are used for $\be$. Since we assumed that $\tb^* \neq \boldsymbol{0}$, without loss of generality, we assume $\beta_1^*\neq 0$ and its absolute value is lower bounded by some constant $c>0$ (i.e., $|\beta^*_1|\geq c$). Let $f$ and $g$ be the density functions of $\X$ when $Y= 1$ and $Y=-1$ respectively. Let $f(x|\X_{-1})$ be the conditional density function of $X_1$ given $(X_2,...,X_p)^\mathrm{T}$ and $f_{-1}(\x_{-1})$ be the joint density of $(X_2,...,X_p)^\mathrm{T}$. Similar notations are used for $g(\cdot)$.

We state some regularity conditions to facilitate theoretical development of asymptotic properties of $\A_{n,h}$ and $\D_{n,h}$.

\begin{enumerate}
	\item[(C0)] There exists a unique nonzero minimizer $\tb^*$ for \eqref{eqn:pop} with $\S(\tb^*) = 0$, and $c\leq \lambda_{\text{min}}(\D(\tb^* ))\leq \lambda_{\text{max}}(\D(\tb^* ))\leq c^{-1}$ for some constant $c>0$.
	\item[(C1)] $|\beta_1^*|\ge c$ and $\lVert\tb ^*\rVert_2\le C$ for some constants $c,C>0$.
	\item[(C2)] Assume that $\sup_{x\in \mathbb{R}}|f(x|\X_{-1})|\le C$, $\sup_{x\in \mathbb{R}}|f'(x|\X_{-1})|\le C$, $\sup_{x\in \mathbb{R}}|xf'(x|\X_{-1})|\le C$, $\sup_{x\in \mathbb{R}}|xf(x|\X_{-1})|\le C$, $\sup_{x\in \mathbb{R}}|x^2f(x|\X_{-1})|\le C$ and $\sup_{x\in \mathbb{R}}|x^2f'(x|\X_{-1})|\le C$ for some constant $C>0$. Also assume $\int_{\mathbb{R}}|x|f(x|\X_{-1})dx<\infty$. Similar assumptions are made for $g(\cdot)$. 
	\item[(C3)] Assume that $p=o(nh/\log n)$ and
	$\sup_{\lVert\v\rVert_2\le1}\mathbb{E}\exp(t_{0}|\v^\mathrm{T}\X|^2)\le C$ for some $t_{0}>0$ and $C>0$.
	\item[(C4)] The smoothing function $H(x)$ satisfies $H(x) = 1$ if $x\geq 1$ and $H(x) = 0 $ if $x\leq -1$, and also assume that $H$ is twice differentiable and $H^{(2)}$ is bounded. Moreover, assume that $h=o(1)$.
\end{enumerate}

As we discussed in Section \ref{sec:pre}, condition (C0) is a standard assumption which can be implied by some mild conditions (see \citealp{koo2008bahadur} (A1)-(A4)). Conditions (C1) is a mild condition on the boundness of $\tb^*$. Condition (C2) is a regularity condition on the conditional density of $f$ and $g$, and it is satisfied by commonly used density functions, e.g., Gaussian distribution and uniform distribution. Condition (C3) is a sub-Gaussian condition on $\X$. Condition (C4) is a smoothness condition on the smooth function $H(\cdot)$ and can be easily satisfied by a properly chosen $H(\cdot)$ (e.g., see an example in Section \ref{sec:sim}).

Under the above conditions, we give Proposition \ref{prop:C} and Proposition \ref{prop:D} for the asymptotic behavior of $\A_{n,h}$ and $\D_{n,h}$, respectively. Recall that $\epsilon_i=1-y_i\tX _i^\mathrm{T}\tb ^*$ and we have the following propositions. The proofs of all results in this section are relegated to Appendix \ref{app:main}.

\begin{proposition}\label{prop:C}
	Under conditions (C0)-(C4), assume that we have an initial estimator $\tb _0$ with $\lVert\tb _0-\tb ^*\rVert_2=O_\mathbb{P}(a_n)$, where $a_n$ is the convergence rate of the initial estimator. We choose the bandwidth such that $a_n=O(h)$, then we have
	\begin{equation*}
	\left\lVert \A_{n,h}(\tb _0)-\frac{1}{n}\sum_{i=1}^ny_i\tX _iI\{\epsilon_i\ge0\}\right\rVert_2=O_\mathbb{P}\left(\sqrt{\frac{ph\log n}{n}}+a_n^2+h^2\right).
	\end{equation*}
\end{proposition}

\begin{proposition}\label{prop:D}
	Suppose the same conditions in Propositions \ref{prop:C} hold, we have
	\begin{equation*}
	\left\lVert \D_{n,h}(\tb _0)-\D(\tb ^*)\right\rVert=O_\mathbb{P}\left(\sqrt{\frac{p\log n}{nh}}+a_n+h\right).
	\end{equation*}
\end{proposition}

According to the above propositions, with some algebraic manipulations and condition (C0), we have
\begin{equation}\label{eqn:onestep}
\tb -\tb ^*=\D(\tb ^*)^{-1}\left(\frac{1}{n}\sum_{i=1}^ny_i\tX _iI\{\epsilon_i\ge0\}-\lambda \binom{0}{\be^{*}}\right)+\r_n,
\end{equation}
with
\begin{equation}\label{eqn:onstepremainder}
\lVert \r_n\rVert_2=O_\mathbb{P}\left(\sqrt{\frac{p^2\log n}{n^2h}}+\sqrt{\frac{ph\log n}{n}}+a_n^2+h^2\right).
\end{equation}

By appropriately choosing the bandwidth $h$ such that it shrinks with $a_n$ at the same rate (see Theorem \ref{thm:bahadur}), $a_n^2$ becomes the dominating term on the right hand side of \eqref{eqn:onstepremainder}. This implies that by taking one round of refinement, the $L_2$ norm of $\tb - \tb^*$ improves from $O_\mathbb{P}(a_n)$ to $O_\mathbb{P}(a_n^2)$ (note that $\|\tb_0-\tb^*\|_2 = O_\mathbb{P}(a_n)$, see Proposition \ref{prop:C}). Therefore by recursively applying the argument in \eqref{eqn:onestep} and setting the obtained estimator as the new initial estimator $\tb_0$, the algorithm iteratively refines the estimator $\tb$.
This gives the Bahadur representation of our MDL estimator $\tb^{(q)}$ for $q$ rounds of refinements (see Algorithm \ref{algo:dsvm}).

\begin{theorem}\label{thm:bahadur}
	Under conditions (C0)-(C4), assume that the initial estimator $\tb _0$ satisfies $\lVert\tb _0-\tb ^*\rVert_2=O_\mathbb{P}(\sqrt{p/m})$. Also, assume $p=O(m/(\log n)^2)$ and $\lambda=O(1/\log n)$. For a given integer $q\ge1$, let the bandwidth in the $g$-th iteration be $$h:=h_g=\max(\lambda,\sqrt{p/n},(p/m)^{2^{g-2}})$$ for $1\le g\le q$. Then we have
	\begin{equation}\label{eqn:bahadur}
	\tb ^{(q)}-\tb ^*=\D(\tb ^*)^{-1}\left(\frac{1}{n}\sum_{i=1}^n y_i\tX _iI\{\epsilon_i\ge0\}-\lambda \binom{0}{\be^{*}}\right)+\r_n,
	\end{equation}
	with
	\begin{equation}\label{eqn:error}
	\lVert\r_n\rVert_2=O_\mathbb{P}\left(\sqrt{\frac{ph_q\log n}{n}}+\left(\frac{p}{m}\right)^{2^{q-1}}+\lambda^{2}\right).
	\end{equation}
\end{theorem}

It is worthwhile noting that the choice of bandwidth $h_g$ in Theorem \ref{thm:bahadur} is up to a constant. One can choose $h_g = C_0\max(\sqrt{p/n},(p/m)^{2^{g-2}})$ for a constant $C_0>0$ in practice and Theorem \ref{thm:bahadur} still holds. We omit the constant  $C_0$ for simplicity of the statement (i.e., setting $C_0=1$). We notice that the algorithm is not sensitive to the choice of $C_0$. Even with a suboptimal constant $C_0$, the algorithm still shows good performance with a few more rounds of iterations (i.e., using a larger $q$).  Please see Section \ref{sec:sim} for a simulation study that shows the insensitivity to the scaling constant.

According to our choice of $h_q$, we can see that as long as the number of iterations satisfies
\begin{equation}\label{eqn:q}
q\ge1+\log_2\left(\frac{\log n-\log p}{\log m- \log p}\right),
\end{equation}
the bandwidth is $h_q = \sqrt{p/n}$. Then by \eqref{eqn:error}, the Bahadur remainder term $\r_n$ becomes
\begin{equation}\label{eqn:baremainder}
\lVert \r_n\rVert_2=O_\mathbb{P}((p/n)^{3/4}(\log n)^{1/2}+\lambda^{2}).
\end{equation}

When $\lambda\geq \sqrt{p/n}$, the convergence rate $\tb ^{(q)}-\tb ^*$ in \eqref{eqn:bahadur} is dominated by $\lambda$. On the other hand, if $\lambda=O(\sqrt{p/n})$, then $\tb ^{(q)}-\tb ^*$ achieves the optimal rate $O_{\mathbb{P}}(\sqrt{p/n})$.

\begin{remark}[The conditions on $p$ and $\lVert\tb _0-\tb ^*\rVert_2$]\label{rmk:init}
	\normalfont
	
	In this paper, we assume that the initializer is computed on the first machine with the convergence rate $\lVert\tb _0-\tb ^*\rVert_2=O_\mathbb{P}(\sqrt{p/m})$. We require $p=O(m/(\log n)^2)$, which not only guarantees the consistency of the estimator, but also provides us with a concise rate in the Bahadur remainder term (see Equation \ref{eqn:error}).
	
	In fact, the assumption is not necessary if we assume that there is an initializer $\tb _0$ that satisfies  $\lVert\tb _0-\tb ^*\rVert_2=O_\mathbb{P}(n^{-\delta})$ for some constant $\delta>0$. Let $h_{g}=\max(\lambda,\sqrt{p/n},n^{-2^{g-1}\delta})$ and assume that conditions (C0)-(C4) hold. By the proof of Theorem \ref{thm:bahadur}, we have
	\begin{eqnarray*}
		\lVert\r_n\rVert_2=O_\mathbb{P}\left(\sqrt{\frac{ph_q\log n}{n}}+n^{-2^{q}\delta}+\lambda^{2}\right).
	\end{eqnarray*}
	As long as the number of iterations $q$ satisfies
	\begin{equation}\label{eq:q2}
	\begin{aligned}
	q\ge\log_2\left(\frac{\log n-\log p}{\delta\log n}\right),
	\end{aligned}
	\end{equation}
	which is usually a small number in practice, we still obtain the optimal rate of the Bahadur remainder term in \eqref{eqn:baremainder}.
\end{remark}

\begin{remark}[Choice of the batch size $m$]
	\label{rmk:batchsize}
	\normalfont
	The data batch size $m$ balances the tradeoff between communication cost and computation cost. More specifically, when the batch size $m$ is large, the convergence rate of the initial estimator is faster.  Then, the  required number of iterations becomes smaller (see Equation \ref{eqn:q}), which leads to a smaller communication cost. On the other hand, for a large batch size $m$, the computation cost of the initial estimator is large. Moreover, the computation time of $\U_k$ and $\V_k$ on each local machine also grows linearly in $m$. When the batch size $m$ is small,  the computation of the initial estimator becomes faster but it requires more iterations  to achieve the same performance.
	
	In practice, when the data is collected by multiple machines, the batch size will naturally be the storage size of each local machine. If we are allowed to specify $m$, we first need to make sure that $m$ should be large enough so that the initial estimator is consistent.  Moreover, since the communication is usually the bottleneck in distributed computing, it is desirable to choose $m$ to be as large as possible to reach the capacity/memory limit of each local machine.
	This will provide a faster convergence rate of the initial estimator, and thus leads to a smaller number of iterations.
\end{remark}
\begin{remark}[Unbalanced batch size case]
	\label{rmk:unbalanced}
	\normalfont
	
	When the sample sizes on local machines are not balanced, we will choose the machine with the largest local sample size as the first machine to compute the initial estimator. This will provide us an initial estimator with faster convergence rate. As compared to the balanced case, our MDL estimator will require a smaller number of iterations to achieve the optimal statistical efficiency.  It is worth noting that after the initial estimator is given, the MDL estimator in Algorithm 1 does not depend on the sample size on each local machine.\end{remark}

Define $\G(\tb ^*)=\mathbb{E}[\tX \tX ^\mathrm{T}I\{1-Y\tX ^\mathrm{T}\tb ^*\ge0\}]$. By applying the central limit theorem to \eqref{eqn:bahadur}, we have the following result on the asymptotic distribution of $\tb ^{(q)}-\tb ^*$.
\begin{theorem}\label{thm:asym}
	Suppose that all the conditions of Theorem \ref{thm:bahadur} hold with $h=h_g$ and $\lambda=o(n^{-1/2})$. Further, assume that $n=O(m^A)$ for some constant $A\ge1$, $p=o(\min\{n^{1/3}/(\log n)^{2/3},m^\nu\})$ for some $0<\nu<1$ and $q$ satisfies \eqref{eqn:q}. For any nonzero $\tilde{\v}\in\mathbb{R}^{p+1}$, we have as $n,p \rightarrow \infty$,
	\begin{equation*}
	\frac{n^{1/2}\tilde{\v}^\mathrm{T}(\tb ^{(q)}-\tb ^*)}{\sqrt{\tilde{\v}^\mathrm{T}\D(\tb ^*)^{-1}\G(\tb ^*)\D(\tb ^*)^{-1}\tilde{\v}}}\rightarrow\mathcal{N}(0,1).
	\end{equation*}
\end{theorem}

Please see Appendix \ref{app:main} for the proofs of Theorem \ref{thm:bahadur} and Theorem \ref{thm:asym}. We impose the conditions  $n = O(m^A)$ and $p= o(m^\nu)$  for some constants $A \geq 1$ and $\nu \in (0,1)$ in order to ensure the right hand side of \eqref{eqn:q} is bounded by a constant,  which implies that we only need to perform a constant number of iterations even when $n,m \rightarrow \infty$.

We introduce the vector $\tilde{\v}$ since we consider the diverging $p$ regime and thus the dimension of  the ``sandwich matrix'' $\D(\tb ^*)^{-1}\G(\tb ^*)\D(\tb ^*)^{-1}$ is growing in $p$. Therefore, it is notationally convenient to introduce an arbitrary vector $\tilde{\v}$ to make the limiting variance $\tilde{\v}^\mathrm{T}\D(\tb ^*)^{-1}\G(\tb ^*)\D(\tb ^*)^{-1}\tilde{\v}$ a positive real number.
Also note that the conditions $p= o(n^{1/3}/(\log n)^{2/3})$  guarantees that the remainder term \eqref{eqn:baremainder} satisfies $\|\r_n\|_2 = o_\mathbb{P}(n^{-1/2})$, which enables the application of the central limit theorem.

It is also important to note  that the asymptotic variance  $\tilde{\v}^\mathrm{T}\D(\tb ^*)^{-1}\G(\tb ^*)\D(\tb ^*)^{-1}\tilde{\v}$ in Theorem \ref{thm:asym} matches the optimal asymptotic variance of $\tb _\text{SVM all}$ in \eqref{eqn:all},  which is directly computed on all samples (see  Theorem 2 in \citealp{koo2008bahadur}). This result shows that the MDL estimator $\tb ^{(q)}$ does not lose any statistical efficiency as compared to the linear SVM in a single machine setup. \label{discussion:communication} By contrast, the na\"ive divide-and-conquer approach requires the number of local machines $N$ to satisfy the condition $N\leq (n/\log(p))^{1/3}$ (see Remark 1 in \citealp{lian2017divide}). When this condition fails, the asymptotic normality of the estimator no longer holds. The MDL estimator removes the restriction on the number of machines but requires more communications overhead. In particular, the total communication cost for the MDL approach is $O(p^2 N q)$ (or $O(p NqT)$ for Algorithm \ref{algo:dsvm_efficient}), as compared to the one-shot communication $O(pN)$ in the na\"ive divide-and-conquer approach. It is worth noting that, under the assumptions $n = O(m^A)$ and $p = O(m^\nu)$ for some constants $A>0$ and $0<\nu<1$ (see Proposition \ref{prop:efficient}), a constant number of iterations (i.e., $qT=O(1)$) is enough to achieve the optimal rate.

We note that to construct the confidence interval of $\tilde{\v}^\mathrm{T}\tb^{*}$ based on Theorem \ref{thm:asym}, we need consistent estimators of $\D(\tb^{*})$ and $\G(\tb^{*})$. Since $\G(\tb^{*})$ is defined as an expectation, it is natural to estimate it by its empirical version $\widehat{\G}(\tb^{(q)})$.
Moreover, by Proposition \ref{prop:D}, we can estimate $\D(\tb^{*})$ by $\D_{n,h}(\tb^{(q-1)}) = N^{-1}\sum_{k=1}^{N}\V_k$. Since  $\D_{n,h}(\tb^{(q-1)})$ has already been obtained in the algorithm in the last iteration, we don't need extra computation.
Given the nominal coverage probability $1-\rho_0$, the confidence interval for $\tilde{\v}^\mathrm{T}\tb^{*}$ is given by
\begin{equation}\label{eqn:ci}
\tilde{\v}^\mathrm{T}{\tb^{(q)}} \pm n^{-1/2}z_{\rho_0/2}\sqrt{\tilde{\v}^\mathrm{T}\widehat{\D}(\tb^{(q-1)})^{-1}\widehat{\G}(\tb^{(q)})\widehat{\D}(\tb^{(q-1)})^{-1}\tilde{\v}},
\end{equation}
where $z_{\rho_0/2}$ is the $1-\rho_0/2$ quantile of the standard normal distribution. For a fixed vector $\tilde{\v} $, denote $\hat{\sigma}_{n,q} =\sqrt{\tilde{\v}^\mathrm{T}\widehat{\D}(\tb^{(q-1)})^{-1}\widehat{\G}(\tb^{(q)})\widehat{\D}(\tb^{(q-1)})^{-1}\tilde{\v}}$. We have the following theorem for the asymptotic validity for the constructed confidence interval.
\begin{theorem}[Plug-in estimation of the confidence interval]\label{col:plugin}
	Under the conditions of Theorem \ref{thm:asym}, for any nonzero $\tilde{\v}\in\mathbb{R}^{p+1}$, we have as $n,p \rightarrow \infty$,
	\[
	\mathbb{P}\left(\tilde{\v}^\mathrm{T}{\tb^{(q)}} - n^{-1/2}z_{\rho_0/2}\hat{\sigma}_{n,q} \leq \tilde{\v}^\mathrm{T}\tb ^* \leq \tilde{\v}^\mathrm{T}{\tb^{(q)}} + n^{-1/2}z_{\rho_0/2}\hat{\sigma}_{n,q}\right) \rightarrow 1-\rho_0.
	\]
\end{theorem}
In the proof, we first show that $\widehat{\G}(\tb^{(q)})$ is a consistent estimator of ${\G}(\tb^{*})$. The detailed proof of Theorem \ref{col:plugin} is relegated to Appendix \ref{app:main}.

\begin{remark}[Kernel SVM]
	\normalfont
	It is worthwhile to note that the proposed distributed algorithm can also be used in solving nonlinear SVM by using feature mapping approximation techniques. In the general SVM formulation, the objective function is defined as follows:
	\begin{equation}\label{eqn:kerprimal}
	\min_{\tb=(\beta_0, \be)} \sum_{i=1}^n \left(1-y_i(\phi(\X_i)^\mathrm{T}\be+\beta_0)\right)_++\frac{\lambda}{2}\|\be\|^2_{2},
	\end{equation}
	where the function $\phi$ is the feature mapping function which maps $\X_i$ to a high or even infinite dimensional space. The function $K:\R^p\times\R^p\rightarrow\R$ defined by $K(\x,\z) = \phi(\x)^\mathrm{T}\phi(\z)$ is called the kernel function associated with the feature mapping $\phi$. With kernel mapping approximation, we construct a low dimensional feature mapping approximation $\psi:\R^p\rightarrow\R^d$ such that $\psi(\x)^\mathrm{T}\psi(\z) \approx  \phi(\x)^\mathrm{T}\phi(\z)= K(\x,\z)$. Then the original nonlinear SVM problem \eqref{eqn:kerprimal} can be approximated by
	\begin{equation}\label{eqn:kerapp}
	\min_{\tb \in \R^{d+1} } \sum_{i=1}^n \left(1-y_i(\psi(\X_i)^\mathrm{T}\be+\beta_0)\right)_++\frac{\lambda}{2}\|\be\|^2_{2}.
	\end{equation}
	Several feature mapping approximation methods have been developed for kernels with some nice properties  (see, e.g., \citealp{rahimi2008random,lee2011approximate,vedaldi2012efficient}), and it is also shown that the approximation error $|\psi(\x)^\mathrm{T}\psi(\z)- K(\x,\z)|$ is small under some regularity conditions. We note that we should use a data-independent feature mapping approximation where $\psi$ only depends on the kernel function $K$. This ensures that $\psi$ can be directly computed without loading data, which enables efficient algorithm in a distributed setting. For instance, for the RBF kernel, which is defined as $K_{RBF}(\x,\z) = \exp\left(-\sigma{\|\x-\z\|_2^2}\right)$, \cite{rahimi2008random} proposed a data-independent approximation $\psi$ as
	\[
	\psi(\X) = \sqrt{\frac{2}{d}} [\cos(\v_1^\mathrm{T}\X+\omega_1),...,\cos(\v_d^\mathrm{T}\X+\omega_d)]^\mathrm{T},
	\]
	where $\v_1,...,\v_d\in \R^p$ are i.i.d. samples from a multivariate Gaussian distribution $\mathcal{N}(\mathbf{0}, 2\sigma \mathbf{I})$ and $\omega_1,...,\omega_d$ are i.i.d. samples from the uniform distribution on $[0,2\pi]$.
\end{remark}

\begin{remark}[High-dimensional extension]
	\label{rmk:high_d}
	\normalfont
	We note that it is possible to extend the proposed MDL estimator to the high-dimensional case. In particular, with the smoothing technique used in our paper, we have the following smoothed loss function,
	\[
	{L_n}(\tb) = \frac{1}{n} \sum_{i=1}^n K_h(1-y_il(\X_i;\tb ))+\frac{\lambda}{2}\lVert\be\rVert^2_{2},
	\]
	where $K_h(u) = uH(\frac{u}{h})$ is the smoothed hinge loss. Since  the loss function ${L_n}(\tb) $ is second-order differentiable, we can adopt the regularized approximate Newton method (see, e.g., \citealp{jordan2016communication,wang2016efficient}).  More specifically, the regularized approximate Newton method considers the following estimator,
	\begin{equation}\label{eqn:AN}
	\begin{aligned}
	\tb_{1} = \argmin_{\tb\in \R^{p+1}} \{{L}_1(\tb)-\tb^{\rm T}(\nabla {L}_1(\tb_0)-\nabla {L_n}(\tb_0))+\frac{\tilde{\lambda}}{2}\lVert\be\rVert_1\},
	\end{aligned}
	\end{equation}
	where $\tb_0$ is an initial estimator and ${L}_1(\tb)$ is the smoothed loss function computed using the data on the first machine. It can be extended to an iterative algorithm by repeatedly updating the estimator using \eqref{eqn:AN}. However, there are two technical challenges. First, the smoothed loss function becomes non-convex. Second, it is unclear how to choose the bandwidth $h$ such that it shrinks properly with the number of iterations. We leave these two technical questions and the extension to the high-dimensional case for future investigation.
\end{remark}

\section{Simulation Studies}
\label{sec:sim}
In this section, we provide a simulation experiment to illustrate the performance of the proposed distributed SVM algorithm. The data is generated from the following model
\[
P(Y_i = 1) = p_+,\quad P(Y_i = -1) = p_-=1-p_+,
\]
\[
\X_i = Y_i\textbf{1}+\eps_i,\quad\eps_i\sim \mathcal{N}(\textbf{0},\sigma^2\textbf{I}),\quad i=1,2,...,n,
\]
where $\textbf{1}$ is the all-one vector $(1,1,...,1)^\mathrm{T}\in \mathbb{R}^p$ and the triplets $(Y_i,\X_i,\eps_i)$ are drawn independently. We set $\sigma =\sqrt{p}$ throughout the simulation study. In order to directly compare the proposed estimator to other estimators, we follow the simulation study setting in \cite{koo2008bahadur} and consider the optimization problem without penalty term, i.e., $\lambda = 0$. We set $p_+=p_-=\frac{1}{2}$, i.e., the data is generated from the two classes with equal probability. Note that we can explicitly solve the true coefficient $\tb^*$ by the following claim whose proof is relegated to Appendix \ref{app:aux}.

\begin{claim}\label{claim:sim}
	The true coefficient vector is $\tb^{*} = \frac{1}{a}(0,1,...,1)^\mathrm{T}\in \mathbb{R}^{p+1}$, where $a$ is the solution to $\int_{-\infty}^{a} \phi_1(x)x dx=0$ and $\phi_1(x)$ is the p.d.f. of the distribution $\mathcal{N}(p,\sigma^2 p)$.
\end{claim}

We use the integral of a kernel function as the smoothing function:
\begin{equation*}
H(v)=\left\{
\begin{array}{ll}
0&\text{if }v\leq -1,\\
\frac{1}{2}+\frac{15}{16}\left(v-\frac23v^3+\frac15 v^5\right)&\text{if }|v| < 1,\\
1&\text{if }v \geq 1.
\end{array}
\right.
\end{equation*}

The initial estimator $\tb_0$ is computed by directly solving the convex optimization problem \eqref{opt} with only the samples in the first machine, and the iterative distributed algorithm is then applied to data in all the machines. We consider the na\"ive divide-and-conquer (Na\"ive-DC) approach which simply computes the solution of the optimization problem on every single machine and combines all the solutions by taking the average. The oracle estimator is defined by \eqref{eqn:all} which directly solves the optimization with data from all machines.
The confidence intervals are constructed for $\tilde{\v}_0^\mathrm{T}\tb^{*}$ with all these three estimators, where $\tilde{\v}_0 = (p+1)^{-1/2}\textbf{1}_{p+1}$ and the nominal coverage probability $1-\rho_0$ is set to 95\%. We use \eqref{eqn:ci} to construct the confidence interval and we also use the same interval length for all the three estimators. We compare both the $L_2$ distance between the estimator and the true coefficient $\tb^{*}$ and the empirical coverage rate for all the three estimators.

\subsection{$L_2$ Error and Empirical Coverage Rate}
\begin{figure}[t!]
	\centering
	\begin{subfigure}{0.49\textwidth} 
		\includegraphics[width=\textwidth]{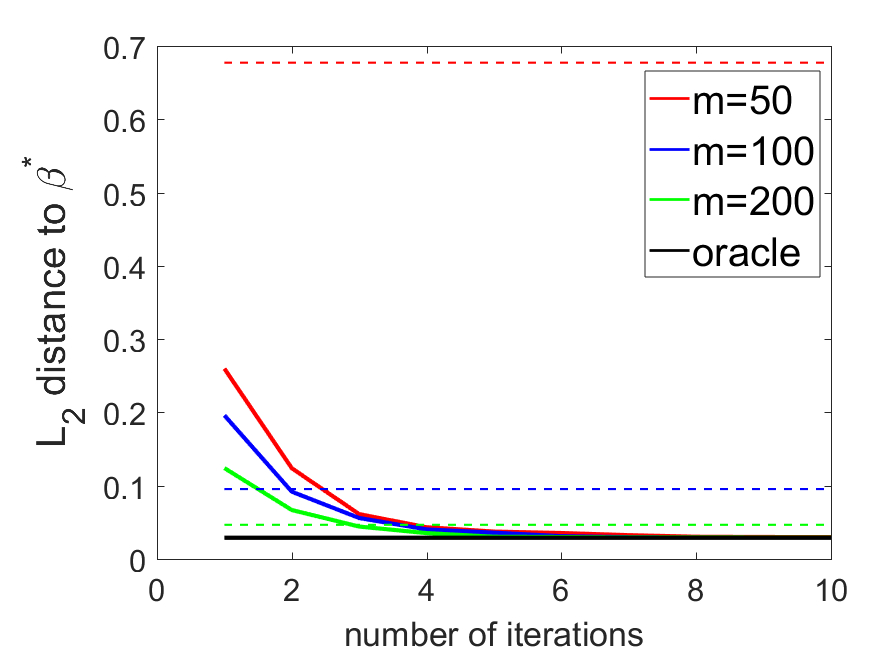}
		\caption{$p=4$, $n=10^4$ case} 
		\label{q1}
	\end{subfigure}
	\vspace{1em} 
	\begin{subfigure}{0.49\textwidth} 
		\includegraphics[width=\textwidth]{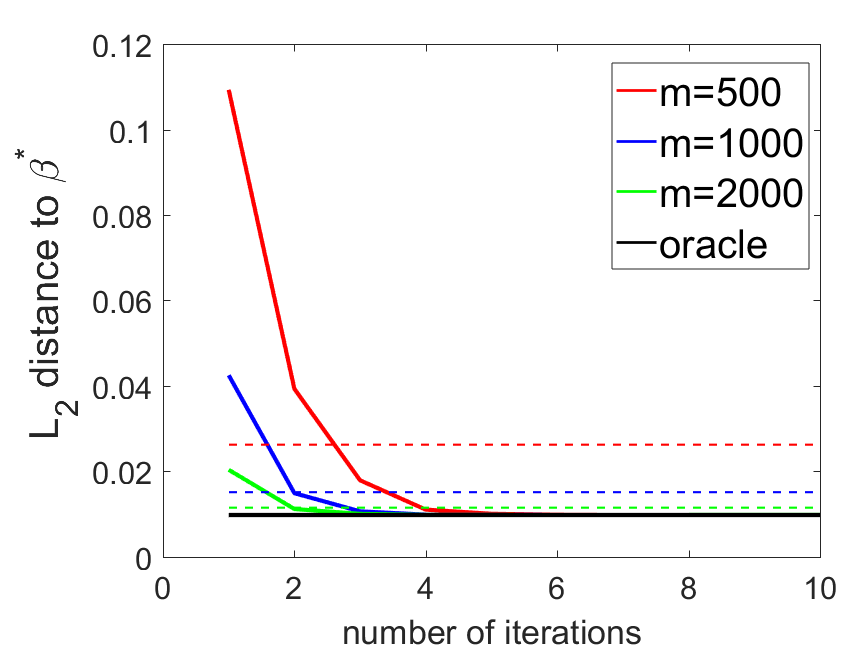}
		\caption{$p=20$, $n=10^6$ case} 
		\label{q2}
	\end{subfigure}
	\caption{$L_2$ error of three estimators with different number of iterations $q$.  The dashed horizontal lines show the performance of the Na\"ive-DC  for different values of $m$ and the solid lines show the performance of the MDL estimators (for different $m$) and the oracle estimator.} 
	\label{fig:q}
\end{figure}
\begin{figure}[t!]
	\centering
	\begin{subfigure}{0.99\textwidth} 
		\includegraphics[width=\textwidth]{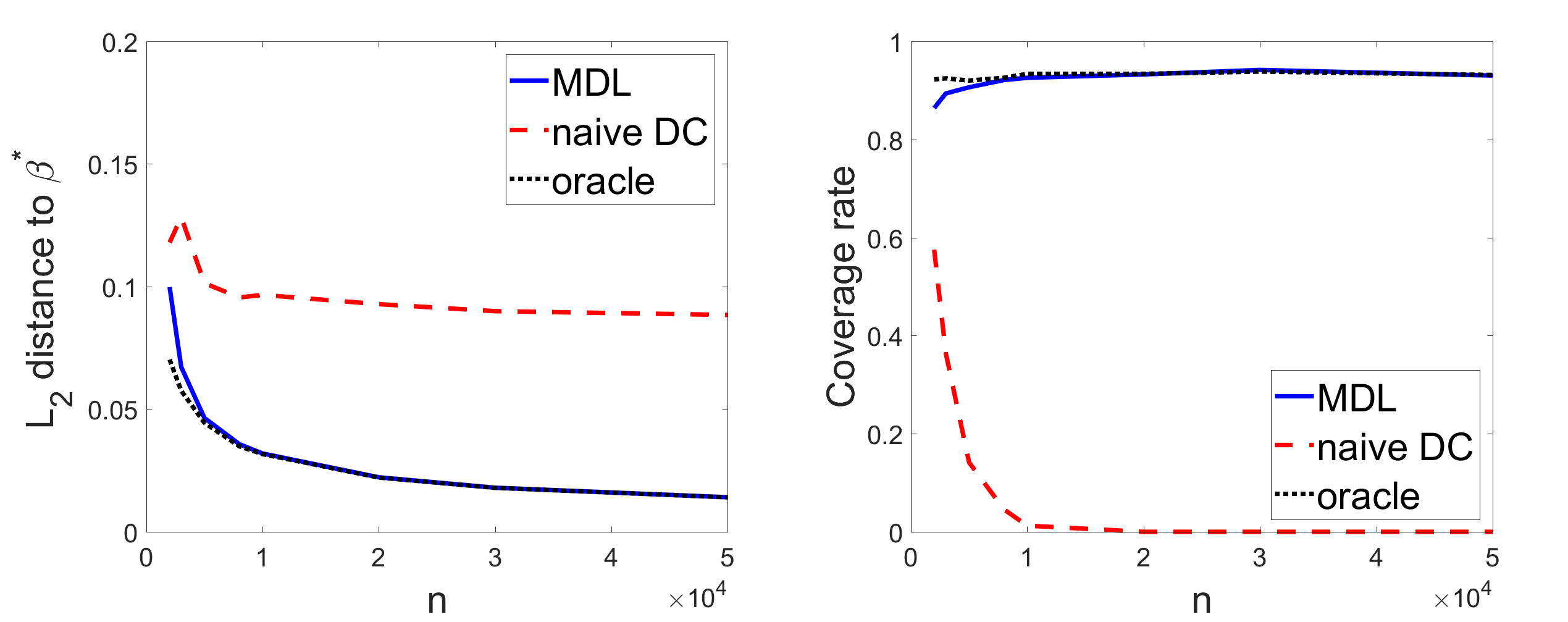}
		\caption{$p=4$, $m=100$ case} 
		\label{n1}
	\end{subfigure}
	\vspace{1em} 
	\begin{subfigure}{0.99\textwidth} 
		\includegraphics[width=\textwidth]{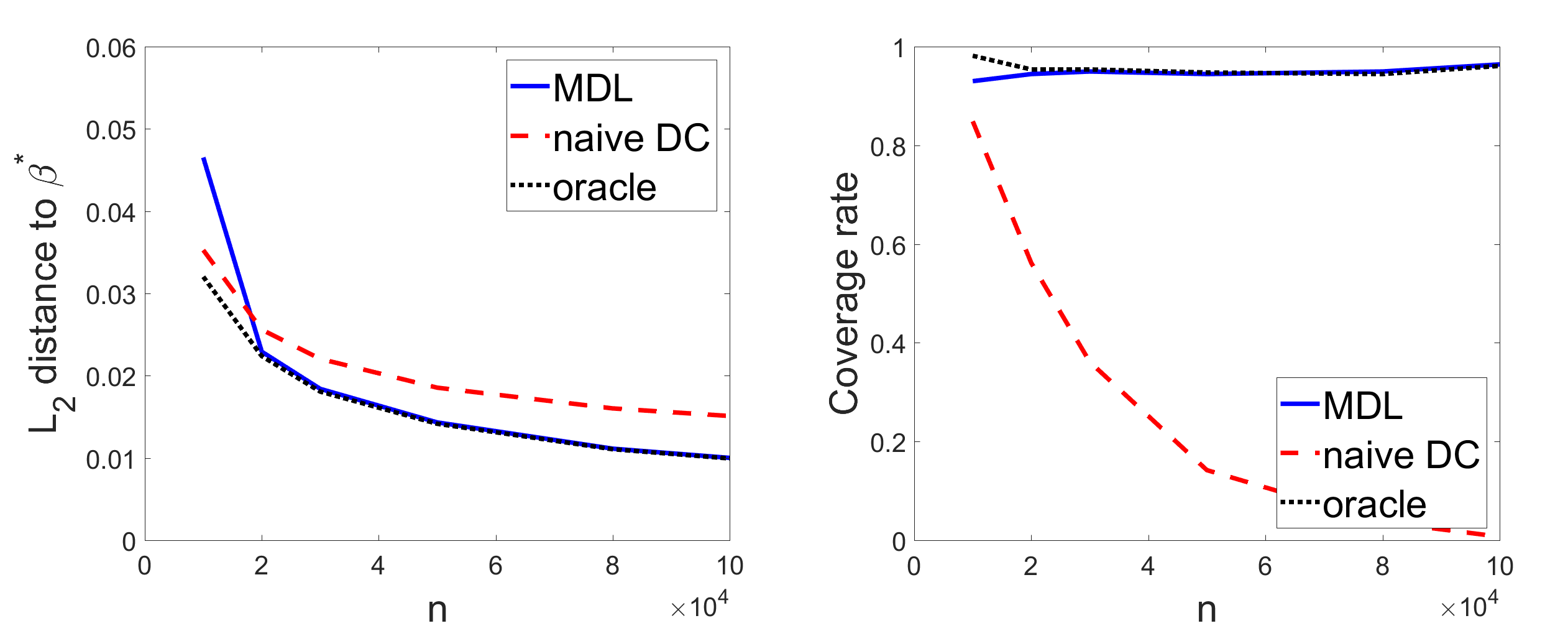}
		\caption{$p=20$, $m=1000$ case} 
		\label{n2}
	\end{subfigure}
	\caption{$L_2$ error and coverage rate of the MDL estimator with different total sample size $n$ (the number of iterations $q=10$).} 
	\label{fig:n}
\end{figure}
We first investigate how the $L_2$ error of our proposed estimator improves with the number of aggregations. We consider two settings: the number of samples $n = 10^4$, dimension $p=4$, batch size $m \in \{50,100,200\}$ and $n = 10^6$, $p=20$, $m \in \{500,1000,2000\}$. We set the max number of iterations as 10 and plot the $L_2$ error at each iteration. We also plot the $L_2$ error of Na\"ive-DC estimator (the dashed line) and the oracle estimator (the black line) as horizontal lines for comparison. All the results reported are the average of 1000 independent runs.
\begin{figure}[t!]
	\centering
	\begin{subfigure}{0.99\textwidth} 
		\includegraphics[width=\textwidth]{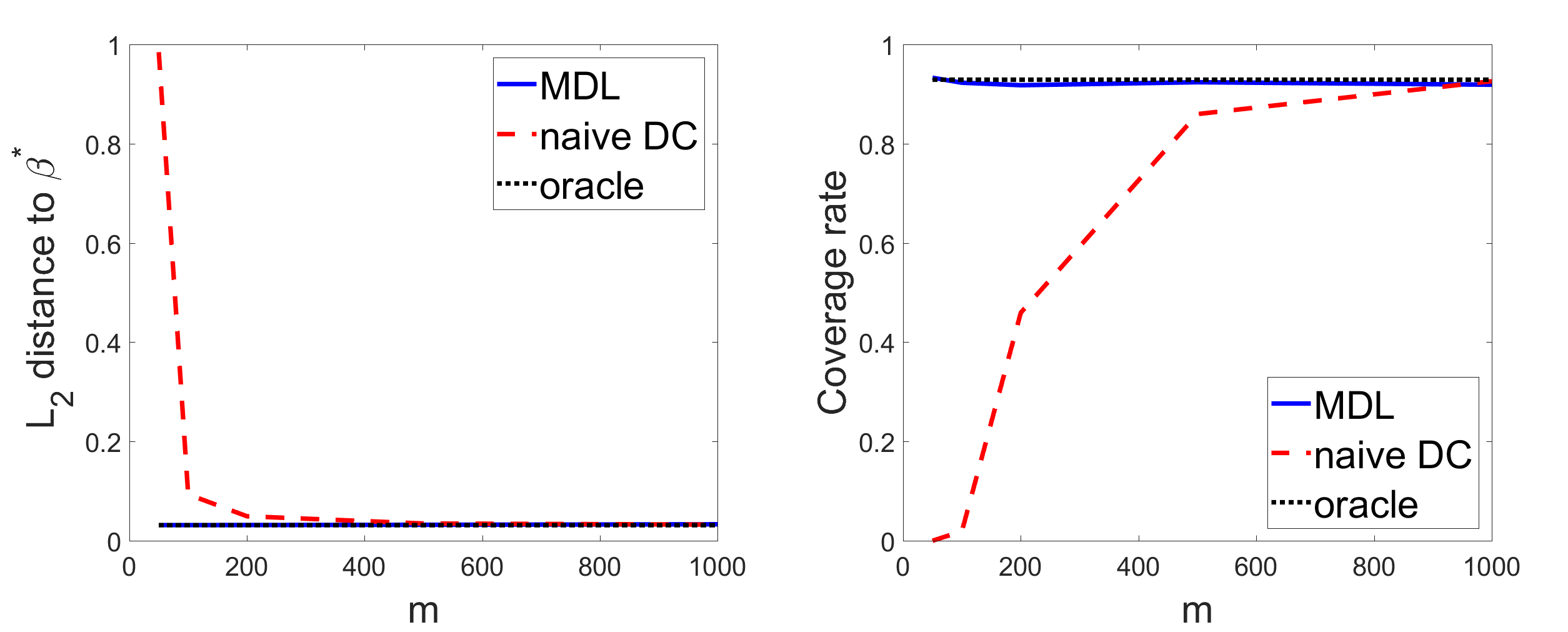}
		\caption{$p=4$, $n=10^5$ case} 
		\label{m1}
	\end{subfigure}
	\vspace{1em} 
	\begin{subfigure}{0.99\textwidth} 
		\includegraphics[width=\textwidth]{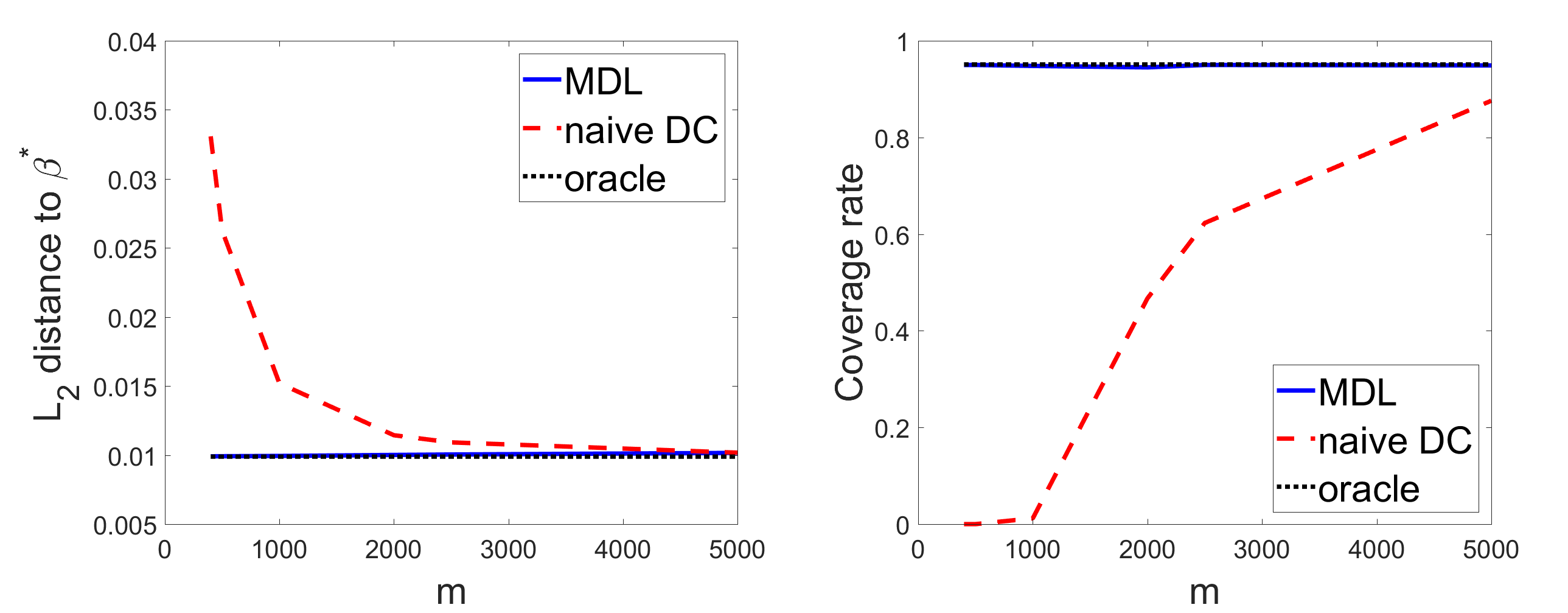}
		\caption{$p=20$, $n=10^6$ case} 
		\label{m2}
	\end{subfigure}
	\caption{$L_2$ error and coverage rate of the MDL estimator with different batch size $m$ (the number of iterations $q=10$).} 
	\label{fig:m}
\end{figure}
From Figure \ref{fig:q} we can see that the error of proposed MDL estimator decreases quickly with the number of iterations. After 5 rounds of aggregations, the MDL estimator performs better than the Na\"ive-DC approach and it almost achieves the same $L_2$ error as the oracle estimator.

Next, we experiment on how the performance of the estimators changes with the total number of data points $n$ while the number of data that each machine can store is fixed. We consider two settings where the machine capacity $m=100$ and $1000$, the number of iterations $q = 10$ and dimension $p = 4$ and $20$, and we plot the $L_2$ error and empirical coverage rate for all the three estimator against the sample size $n$.

From Figure \ref{fig:n} we can observe that the $L_2$ error of the oracle estimator decreases as $n$ increases, but the Na\"ive-DC estimator clearly fails to converge to the true estimator which is essentially due to the fact that the bias of the Na\"ive-DC estimator does not decrease with $n$. However, the proposed MDL estimator converges to the true coefficient with almost the identical rate as the oracle estimator. We also notice that the coverage rate of the MDL estimator is quite close to that of the oracle estimator which is close to the nominal coverage probability 95\%, while the coverage rate of the Na\"ive-DC estimator quickly decreases and drops to zero when $n$ increases.

The next experiment shows how the $L_2$ error and the coverage rate change with different machine capacity $m$ with fixed sample size $n$. Two parameter settings are considered where the sample size $n = 10^5,10^6$, dimension $p=4,20$, and the number of iterations is $q=10$. The results are shown in Figure \ref{fig:m}. From Figure \ref{fig:m} we can see that when the machine capacity gets small, the $L_2$ error of the Na\"ive-DC estimator increases drastically and it fails when $m\leq 100$ in the $n=10^5$ case and $m \leq 400$ in the $n=10^6$ case. On the contrary, the MDL estimator is quite robust even when the machine capacity is small. Moreover, the empirical coverage rate for the Na\"ive-DC estimator is small and only approaches 95\% when $m$ is sufficiently large, while the coverage rate for the proposed MDL estimator is close to the oracle estimator which is close to the nominal coverage probability 95\%.

\subsection{Bias and Variance Analysis}
\begin{table}[t!]
	\centering
	\renewcommand{\arraystretch}{1.4}
	\begin{tabular}{c|c|cc|cc|cc}
		\hline
		\multirow{2}{*}{$(n,p)$} & \multirow{2}{*}{$m$} & \multicolumn{2}{c|}{MDL} & \multicolumn{2}{c|}{Na\"ive-DC} & \multicolumn{2}{c}{Oracle} \\ \cline{3-8}
		&  & \begin{tabular}[c]{@{}c@{}}$\text{bias}^2$\\ $(\times 10^{-4})$\end{tabular} & \begin{tabular}[c]{@{}c@{}}var\\ $(\times 10^{-4})$\end{tabular} & \begin{tabular}[c]{@{}c@{}}$\text{bias}^2$\\ $(\times 10^{-4})$\end{tabular} & \begin{tabular}[c]{@{}c@{}}var\\ $(\times 10^{-4})$\end{tabular} & \begin{tabular}[c]{@{}c@{}}$\text{bias}^2$\\ $(\times 10^{-4})$\end{tabular} & \begin{tabular}[c]{@{}c@{}}var\\ $(\times 10^{-4})$\end{tabular} \\ \hline
		\multirow{4}{*}{$(10^4,4)$} & 100 & 0.004 & 3.906 & 59.329 & 4.457 & 0.000 & 2.275 \\
		& 200 & 0.002 & 2.516 & 10.678 & 2.950 & 0.000 & 2.275 \\
		& 500 & 0.005 & 2.459 & 1.393 & 2.581 & 0.000 & 2.275 \\
		& 1000 & 0.006 & 2.608 & 0.304 & 2.420 & 0.000 & 2.275 \\ \hline
		\multirow{6}{*}{$(10^5,20)$} & 400 & 0.000 & 0.076 & 9.759 & 0.085 & 0.000 & 0.058 \\
		& 500 & 0.000 & 0.059 & 5.351 & 0.080 & 0.000 & 0.058 \\
		& 1000 & 0.000 & 0.059 & 1.140 & 0.069 & 0.000 & 0.058 \\
		& 2000 & 0.000 & 0.060 & 0.261 & 0.063 & 0.000 & 0.058 \\
		& 2500 & 0.000 & 0.060 & 0.168 & 0.062 & 0.000 & 0.058 \\
		& 5000 & 0.000 & 0.061 & 0.041 & 0.058 & 0.000 & 0.058 \\ \hline
	\end{tabular}
	\caption{Bias and variance analysis of MDL, Na\"ive-DC and oracle estimator with different batch size $m$ when number of aggregations $q = 6$.}\label{tab:BVm}
	\renewcommand{\arraystretch}{1}
\end{table}
\begin{table}[t!]
	\centering
	\renewcommand{\arraystretch}{1.4}
	\begin{tabular}{c|c|cc|cc|cc}
		\hline
		\multirow{2}{*}{$(m,p)$} & \multirow{2}{*}{\begin{tabular}[c]{@{}c@{}}$n$\\ $(\times 10^3)$\end{tabular}} & \multicolumn{2}{c|}{MDL} & \multicolumn{2}{c|}{Na\"ive-DC} & \multicolumn{2}{c}{Oracle} \\ \cline{3-8}
		&  & \begin{tabular}[c]{@{}c@{}}$\text{bias}^2$\\ $(\times 10^{-4})$\end{tabular} & \begin{tabular}[c]{@{}c@{}}var\\ $(\times 10^{-4})$\end{tabular} & \begin{tabular}[c]{@{}c@{}}$\text{bias}^2$\\ $(\times 10^{-4})$\end{tabular} & \begin{tabular}[c]{@{}c@{}}var\\ $(\times 10^{-4})$\end{tabular} & \begin{tabular}[c]{@{}c@{}}$\text{bias}^2$\\ $(\times 10^{-4})$\end{tabular} & \begin{tabular}[c]{@{}c@{}}var\\ $(\times 10^{-4})$\end{tabular} \\ \hline
		\multirow{7}{*}{(100,4)} & 2 & 1.782 & 55.412 & 58.377 & 24.912 & 0.017 & 13.382 \\
		& 3 & 0.221 & 28.275 & 60.877 & 17.707 & 0.046 & 8.557 \\
		& 5 & 0.120 & 12.694 & 60.696 & 10.267 & 0.020 & 5.016 \\
		& 8 & 0.005 & 4.056 & 58.806 & 6.213 & 0.028 & 3.244 \\
		& 10 & 0.008 & 3.054 & 62.461 & 4.919 & 0.025 & 2.568 \\
		& 20 & 0.006 & 1.400 & 61.609 & 2.589 & 0.005 & 1.325 \\
		& 30 & 0.000 & 1.611 & 60.540 & 1.754 & 0.002 & 0.773 \\ \hline
		\multirow{5}{*}{(1000,20)} & 20 & 0.002 & 0.355 & 1.075 & 0.334 & 0.001 & 0.294 \\
		& 30 & 0.001 & 0.207 & 1.148 & 0.236 & 0.001 & 0.194 \\
		& 50 & 0.001 & 0.122 & 1.147 & 0.144 & 0.001 & 0.121 \\
		& 80 & 0.000 & 0.072 & 1.118 & 0.082 & 0.000 & 0.072 \\
		& 100 & 0.000 & 0.054 & 1.090 & 0.060 & 0.000 & 0.053 \\ \hline
	\end{tabular}
	\caption{Bias and variance analysis of MDL, Na\"ive-DC and oracle estimator with different sample size $n$ when number of aggregations $q = 6$.}\label{tab:BVn}
	\renewcommand{\arraystretch}{1}
\end{table}

In Table \ref{tab:BVm} and Table \ref{tab:BVn}, we report the bias and variance analysis for the MDL, Na\"ive-DC and oracle estimator. In Table \ref{tab:BVm}, we fix two settings of sample size $n$ and dimension $p$ and investigate how the bias and variance of $\tilde{\v}_0^\mathrm{T} \tb$ change with the batch size $m$ for each estimator. As we can see from Table \ref{tab:BVm}, the variance of both the MDL and Na\"ive-DC estimators  is close to the oracle estimator. However, when the batch size $m$ gets relatively small, the bias term of the Na\"ive-DC estimator goes large, and the squared bias quickly exceeds the variance term, which aligns with the discussion in Section \ref{sec:related}. On the other hand, the bias of the MDL estimator stays small and is quite close to the bias of the oracle estimator.

Similarly, in Table \ref{tab:BVn} we fix two settings of $m$ and $p$ and vary the sample size $n$. We observe that the variance of all the three estimators reduces as the sample size $n$ grows large. However, in both settings the squared bias of the Na\"ive-DC estimator does not improve as $n$ increases which also illustrates why the central limit theorem fails for the Na\"ive-DC estimator. On the other hand, the squared bias of the MDL estimator is close to that of the oracle estimator as $n$ gets large.

\subsection{The Performance under Large $n$ and $p$}\label{sec:largescale}
\begin{table}[!t]
	
	\begin{center}
		\renewcommand{\arraystretch}{1.4}
		\begin{tabular}{l|l|lllll}
			\hline
			\multicolumn{1}{c|}{\multirow{2}{*}{\begin{tabular}[c]{@{}c@{}}$(n,p)$\end{tabular}}} &\multicolumn{1}{c|}{\multirow{2}{*}{\begin{tabular}[c]{@{}c@{}}$m$\end{tabular}}} & \multicolumn{5}{c}{MDL ($\times 10^{-2}$)}                                                                                                                                                             \\ \cline{3-7}
			\multicolumn{1}{c|}{}&\multicolumn{1}{c|}{}                                                                               & \multicolumn{1}{c}{\begin{tabular}[c]{@{}c@{}}$q=2$\end{tabular}} & \multicolumn{1}{c}{\begin{tabular}[c]{@{}c@{}} $q=4$\end{tabular}} & \multicolumn{1}{c}{\begin{tabular}[c]{@{}c@{}}$q=6$\end{tabular}} & \multicolumn{1}{c}{\begin{tabular}[c]{@{}c@{}} $q=8$\end{tabular}} & \multicolumn{1}{c}{\begin{tabular}[c]{@{}c@{}}$q=10$\end{tabular}}  \\ \hline
			\multicolumn{1}{c|}{\multirow{3}{*}{\begin{tabular}[c]{@{}c@{}}$(10^6,50)$\end{tabular}}}&1000&3.223&0.313&0.294&0.293&0.293\\
			&2000&0.601&0.290&0.291&0.291&0.291\\
			&2500&0.436&0.292&0.291&0.291&0.291\\\hline
			\multicolumn{1}{c|}{\multirow{3}{*}{\begin{tabular}[c]{@{}c@{}}$(10^6,100)$\end{tabular}}}&2000&2.330&0.372&0.337&0.338&0.338\\
			&2500&1.473&0.339&0.338&0.338&0.338\\
			&5000&0.436&0.340&0.338&0.338&0.338\\\hline
			\multicolumn{1}{c|}{\multirow{3}{*}{\begin{tabular}[c]{@{}c@{}}$(10^6,200)$\end{tabular}}}&2500&3.182&0.540&0.342&0.324&0.324\\
			&4000&1.139&0.351&0.320&0.324&0.324\\
			&5000&0.865&0.336&0.324&0.324&0.324\\\hline
			\multicolumn{1}{c|}{\multirow{3}{*}{\begin{tabular}[c]{@{}c@{}}$(10^6,500)$\end{tabular}}}&6250&3.502&0.823&0.315&0.324&0.320\\
			&8000&1.547&0.309&0.332&0.321&0.321\\
			&10000&1.342&0.319&0.312&0.320&0.321\\\hline
		\end{tabular}
		\renewcommand{\arraystretch}{1}
		\caption{Comparison of the $L_2$ error under different dimensionality $p$ and batch size $m$. The sample size is fixed to $n = 10^6$, and the number of iterations $q=10$.}\label{tab:largep}
	\end{center}
\end{table}
In this section, we investigate the performance of the MDL estimator for varying dimension $p$. In Table \ref{tab:largep}, we choose a large sample size $n  =10^6$, and vary the dimension $p$ and the batch size $m$. We report the $L_2$ error of the MDL estimator with different number of iterations. From the result we can see that our proposed estimator maintains good performance under large scale settings. The $L_2$ error of the MDL estimator becomes small in all settings and stays stable when the number of iterations $q$ is slightly larger (e.g., $q \geq 4$).

\subsection{Sensitivity Analysis of the Bandwidth Constant $C_0$}
\begin{figure}[!t]
	\centering
	\begin{subfigure}{0.49\textwidth} 
		\includegraphics[width=\textwidth]{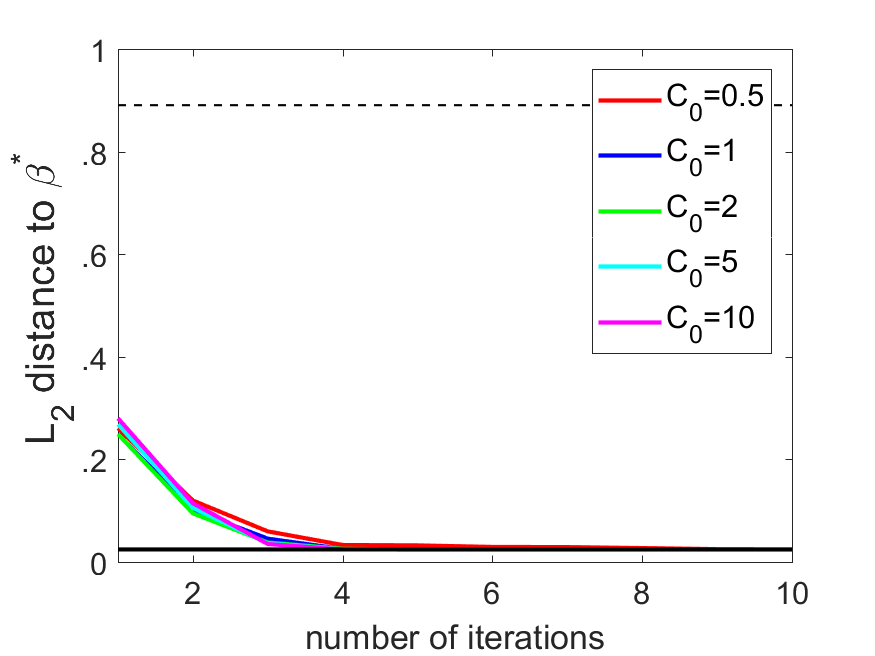}
		\caption{$p=4$, $n=10^4$, $m=50$ case} 
	\end{subfigure}
	\vspace{1em} 
	\begin{subfigure}{0.49\textwidth} 
		\includegraphics[width=\textwidth]{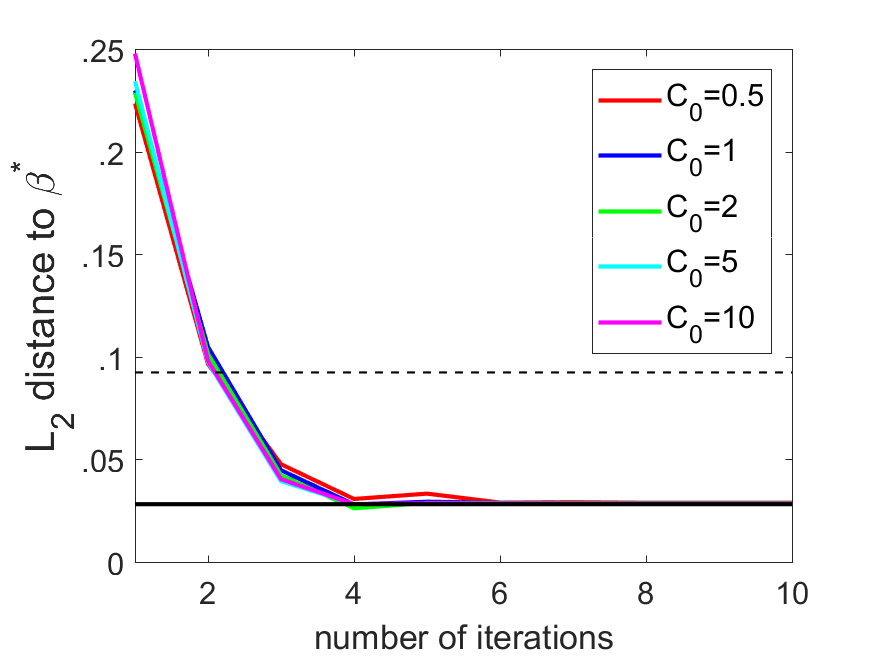}
		\caption{$p=4$, $n=10^4$, $m=100$ case} 
	\end{subfigure}
	\vspace{1em} 
	\begin{subfigure}{0.49\textwidth} 
		\includegraphics[width=\textwidth]{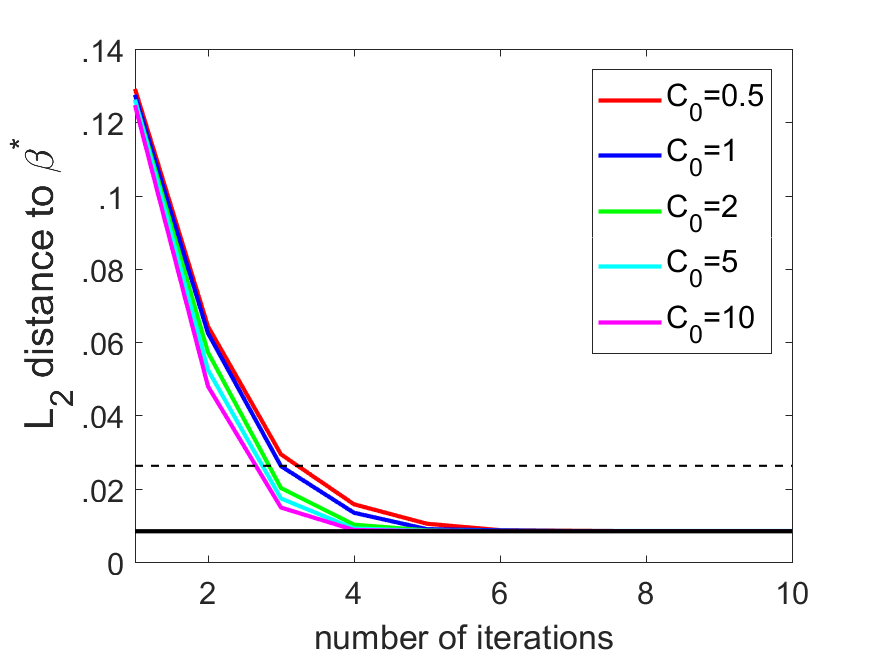}
		\caption{$p=20$, $n=10^5$, $m=500$ case} 
	\end{subfigure}
	\vspace{1em} 
	\begin{subfigure}{0.49\textwidth} 
		\includegraphics[width=\textwidth]{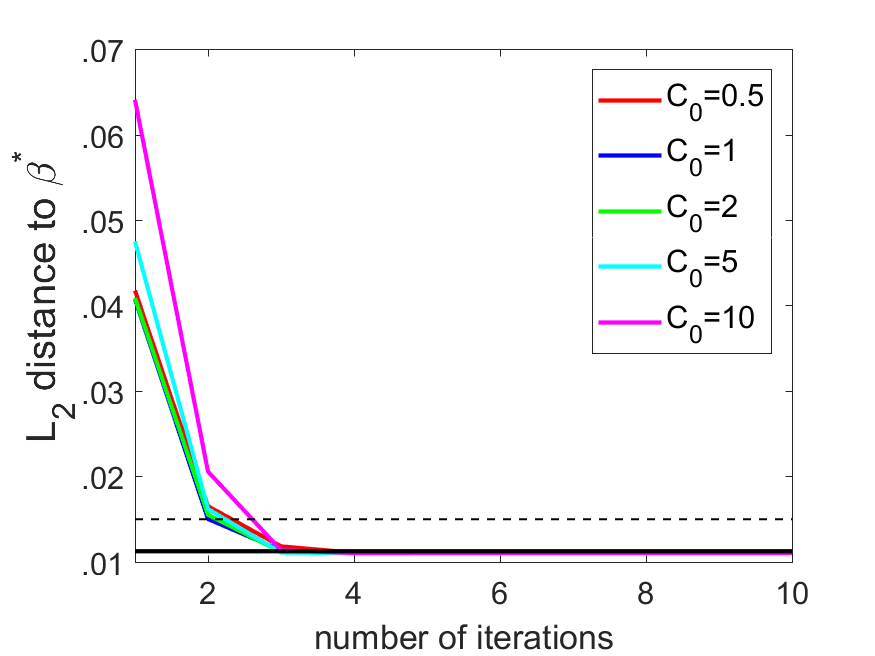}
		\caption{$p=20$, $n=10^5$, $m=1000$ case} 
	\end{subfigure}
	\caption{$L_2$ error of the MDL estimator with different $C_0$. Dashed lines show the performance of the Na\"ive-DC approach and the black solid line shows the performance of the oracle estimator. Other colored lines show the performance of the MDL estimator with different choices of constants in the bandwidth.} 
	\label{fig:C}
\end{figure}
Finally, we report the simulation study to show that the algorithm is not sensitive to the choice of $C_0$ in bandwidth $h_g$ where $h_g = C_0\max(\sqrt{p/n},(p/m)^{2^{g-2}})$. We set $n=10^4$, $p=4$ with $m\in\{50,100\}$ and $n=10^5$, $p=20$ with $m\in\{500,1000\}$. The constant $C_0$ is selected from $\{0.5, 1, 2, 5, 10\}$. We plot the $L_2$ error of the MDL estimator at each iteration step with different choices of $C_0$. We also plot the $L_2$ error of the Na\"ive-DC estimator (the dashed line) and the oracle estimator (the black line) as horizontal lines for comparison. Figure \ref{fig:C} shows that the proposed estimator exhibits good performance for all choices of $C_0$ after a few rounds of iterations and finally achieves the $L_2$ errors which are close to the $L_2$ error of the oracle estimator.

\section{Conclusions and Future Works}
\label{sec:conclusion}
In this paper, we propose a multi-round distributed linear-type (MDL) estimator for conducting inference for linear support vector machine with a large sample size $n$ and a growing dimension $p$. The proposed method only needs to calculate the SVM estimator on a small batch of data as an initial estimator, and all the remaining works are simple matrix operations. Our approach is not only computationally efficient but also achieves the same statistical efficiency as the classical linear SVM estimator using all the data. In our theoretical results in Theorem \ref{thm:bahadur}, the term $(\frac{p}{m})^{2^{q-1}}$ corresponds to the convergence rate of the bias. An interesting theoretical open problem is that whether the rate of the bias is optimal. Note that according to Lemma \ref{lem:1}, the expectation of the bias is bounded by $a_n^2$ (with the choice of bandwidth $h = a_n$). We conjecture that the rate of the bias $a_n^2$ is optimal, but we leave this conjecture for future investigation. 

This work only serves as the first step towards distributed inference for SVM, which is an important area that bridges statistics and machine learning. In the future, we would like to further establish unified computational approaches and theoretical tools for statistical inference for other types of SVM problems, such as $L_q$-penalized SVM (see ,e.g., \cite{liu2007support}), high-dimensional SVM (see, \cite{peng2016error,zhang2016variable}), and more general kernel-based SVM.

\section*{Acknowledgments}
Xiaozhou Wang and Weidong Liu are supported by NSFC, Grant No. 11825104, 11431006 and 11690013, the Program for Professor of Special Appointment (Eastern Scholar) at Shanghai Institutions of Higher Learning, Youth Talent Support Program, 973 Program (2015CB856004), and a grant from Australian Research Council. Zhuoyi Yang and Xi Chen are supported by NSF Award (IIS-1845444), Alibaba Innovation Research Award, and Bloomberg Data Science Research Grant.

\newpage

\appendix
\section{Proofs for Results}
\label{app:theorem}
In this appendix, we provide the proofs of the results.
\subsection{Technical Lemmas}
\label{app:lemma}
Before proving the theorems and propositions, we first introduce three technical lemmas, which will be used in our proof. 

\begin{lemma}\label{lem:1}
	Suppose that conditions (C0)-(C4) hold. For any $\tilde{\v}\in\mathbb{R}^{p+1}$ with $\lVert\tilde{\v}\rVert_2=1$, we have
	\begin{equation*}
	\mathbb{E}\left\{Y\tilde{\v}^\mathrm{T}\tX \left(H\left(\frac{1-Y\tX ^\mathrm{T}\ta }{h}\right)+\frac{1-Y\tX ^\mathrm{T}\tb^{*} }{h}H'\left(\frac{1-Y\tX ^\mathrm{T}\ta }{h}\right)\right)\right\}=O(h^2+\lVert\ta -\tb^{*} \rVert_2^2),
	\end{equation*}
	uniformly in $\lVert\ta -\tb^{*} \rVert_2\le a_n$ with any $a_n\rightarrow0$.
\end{lemma}
\noindent
\textbf{Proof of Lemma \ref{lem:1}.} Without loss of generality, assume that $\beta^{*}_1\ge c$. Then $\alpha_1\ge c/2$. For any $\v\in\mathbb{R}^p$,
\begin{equation*}
\begin{aligned}
&\mathbb{E}\left\{Y(v_0+\v^\mathrm{T}\X)H\left(\frac{1-Y(\alpha_0+\X^\mathrm{T}\al)}{h}\right)\right\}\\
=&\pi_+\int_{\mathbb{R}^p}(v_0+\v^\mathrm{T}\x)H\left(\frac{1-\alpha_0-\x^\mathrm{T}\al}{h}\right)f(\x)d\x-\pi_-\int_{\mathbb{R}^p}(v_0+\v^\mathrm{T}\x)H\left(\frac{1+\alpha_0+\x^\mathrm{T}\al}{h}\right)g(\x)d\x.
\end{aligned}
\end{equation*}
We have
\begin{equation*}
\begin{aligned}
&\int_{\mathbb{R}^p}(v_0+\v^\mathrm{T}\x)H\left(\frac{1-\alpha_0-\x^\mathrm{T}\al}{h}\right)f(\x)d\x\\
=&\int_{\mathbb{R}^{p-1}}\int_{\mathbb{R}}(v_0+v_1x_1+\v_{-1}^\mathrm{T}\x_{-1})H\left(\frac{1-\alpha_0-x_1\alpha_1-\x_{-1}^\mathrm{T}\al_{-1}}{h}\right)f(x_1,\x_{-1})dx_1d\x_{-1}\\
=&-\frac{h}{\alpha_1}\int_{\mathbb{R}^{p-1}}f_{-1}(\x_{-1})\int_{\mathbb{R}}\left(v_0+v_1\frac{1-\alpha_0-\x_{-1}^\mathrm{T}\al_{-1}-hy}{\alpha_1}+\v_{-1}^\mathrm{T}\x_{-1}\right)\\
&\times f\left(\frac{1-\alpha_0-\x_{-1}^\mathrm{T}\al_{-1}-hy}{\alpha_1}|\x_{-1}\right)H(y)dyd\x_{-1}.
\end{aligned}
\end{equation*}
Define $G(t|\x_{-1})=\int_{-\infty}^t xf(x|\x_{-1})dx$. Since $\int_{\mathbb{R}}|x|f(x|\x_{-1})dx<\infty$ , we have $G(-\infty|\x_{-1})=0$. Then,
\begin{equation*}
\begin{aligned}
&-\int_{\mathbb{R}}v_1\frac{1-\alpha_0-\x_{-1}^\mathrm{T}\al_{-1}-hy}{\alpha_1}f\left(\frac{1-\alpha_0-\x_{-1}^\mathrm{T}\al_{-1}-hy}{\alpha_1}|\x_{-1}\right)H(y)dy\\
=&\frac{\alpha_1}{h}v_1\int_{\mathbb{R}}H(y)dG\left(\frac{1-\alpha_0-\x_{-1}^\mathrm{T}\al_{-1}-hy}{\alpha_1}|\x_{-1}\right)\\
=&-\frac{\alpha_1}{h}v_1\int_{-1}^1G\left(\frac{1-\alpha_0-\x_{-1}^\mathrm{T}\al_{-1}-hy}{\alpha_1}|\x_{-1}\right)H'(y)dy\\
=&-\frac{\alpha_1}{h}v_1G\left(\frac{1-\beta^{*}_0-\x_{-1}^\mathrm{T}\be^{*}_{-1}}{\beta^{*}_1}|\x_{-1}\right)\\
&-\frac{\alpha_1}{h}v_1\int_{-1}^1\frac{1-\beta^{*}_0-\x_{-1}^\mathrm{T}\be^{*}_{-1}}{\beta^{*}_1}f\left(\frac{1-\beta^{*}_0-\x_{-1}^\mathrm{T}\be^{*}_{-1}}{\beta^{*}_1}|\x_{-1}\right)\Delta(\al,\be^{*},\x_{-1},y)H'(y)dy\\
&+O(1)\frac{\alpha_1}{h}v_1\int_{-1}^1\Delta^2(\al,\be^{*},\x_{-1},y)|H'(y)|dy,
\end{aligned}
\end{equation*}
where
\begin{equation*}
\Delta(\al,\be^{*},\x_{-1},y)=\frac{1-\alpha_0-\x_{-1}^\mathrm{T}\al_{-1}-hy}{\alpha_1}-\frac{1-\beta^{*}_0-\x_{-1}^\mathrm{T}\be^{*}_{-1}}{\beta^{*}_1},
\end{equation*}
and the inequality $|xf(x|\x_{-1})-yf(y|\x_{-1})|\le C|x-y|$ followed from Condition (C2). Also,
\begin{equation*}
\begin{aligned}
&-\int_{\mathbb{R}}(v_0+\v_{-1}^\mathrm{T}\x_{-1})f\left(\frac{1-\alpha_0-\x_{-1}^\mathrm{T}\al_{-1}-hy}{\alpha_1}|\x_{-1}\right)H(y)dy\\
=&-\frac{\alpha_1}{h}\int_{-1}^1(v_0+\v_{-1}^\mathrm{T}\x_{-1})F\left(\frac{1-\alpha_0-\x_{-1}^\mathrm{T}\al_{-1}-hy}{\alpha_1}|\x_{-1}\right)H'(y)dy\\
=&-\frac{\alpha_1}{h}(v_0+\v_{-1}^\mathrm{T}\x_{-1})F\left(\frac{1-\beta^{*}_0-\x_{-1}^\mathrm{T}\be^{*}_{-1}}{\beta^{*}_1}|\x_{-1}\right)\\
&-\frac{\alpha_1}{h}\int_{-1}^1(v_0+\v_{-1}^\mathrm{T}\x_{-1})f\left(\frac{1-\beta^{*}_0-\x_{-1}^\mathrm{T}\be^{*}_{-1}}{\beta^{*}_1}|\x_{-1}\right)\Delta(\al,\be^{*},\x_{-1},y)H'(y)dy\\
&+O(1)\frac{\alpha_1}{h}\int_{-1}^1|v_0+\v_{-1}^\mathrm{T}\x_{-1}|\Delta^2(\al,\be^{*},\x_{-1},y)|H'(y)|dy.
\end{aligned}
\end{equation*}
Next we consider
\begin{equation*}
\begin{aligned}
&\mathbb{E}\left\{Y(v_0+\v^\mathrm{T}\X)\frac{1-Y(\beta^{*}_0+\X^\mathrm{T}\be^{*})}{h}H'\left(\frac{1-Y(\alpha_0+\X^\mathrm{T}\al)}{h}\right)\right\}\\
=&\pi_+\int_{\mathbb{R}^p}(v_0+\v^\mathrm{T}\x)\frac{1-\beta^{*}_0-\x^\mathrm{T}\be^{*}}{h}H'\left(\frac{1-\alpha_0-\x^\mathrm{T}\al}{h}\right)f(\x)d\x\\
&-\pi_-\int_{\mathbb{R}^p}(v_0+\v^\mathrm{T}\x)\frac{1+\beta^{*}_0+\x^\mathrm{T}\be^{*}}{h}H'\left(\frac{1+\alpha_0+\x^\mathrm{T}\al}{h}\right)g(\x)d\x.
\end{aligned}
\end{equation*}
We have
\begin{equation*}
\begin{aligned}
&\int_{\mathbb{R}^p}(v_0+\v^\mathrm{T}\x)\frac{1-\beta_0^*-\x^\mathrm{T}\be^{*}}{h}H'\left(\frac{1-\alpha_0-\x^\mathrm{T}\al}{h}\right)f(\x)d\x\\
=&-\frac{1}{\alpha_1}\int_{\mathbb{R}^{p-1}}f_{-1}(\x_{-1})\int_{\mathbb{R}}\left(v_0+v_1\frac{1-\alpha_0-\x_{-1}^\mathrm{T}\al_{-1}-hy}{\alpha_1}+\v_{-1}^\mathrm{T}\x_{-1}\right)\\
&\times \left(1-\beta^{*}_0-\beta^{*}_1\frac{1-\alpha_0-\x_{-1}^\mathrm{T}\al_{-1}-hy}{\alpha_1}-\x_{-1}^\mathrm{T}\be^{*}_{-1}\right)\\
&\times f\left(\frac{1-\alpha_0-\x_{-1}^\mathrm{T}\al_{-1}-hy}{\alpha_1}|\x_{-1}\right)H'(y)dyd\x_{-1}.
\end{aligned}
\end{equation*}
Note that

\begingroup
\allowdisplaybreaks
\makeatletter\def\f@size{10}\check@mathfonts
\begin{align*}
&-\frac{1}{\alpha_1}\int_{\mathbb{R}}\left(v_0+v_1\frac{1-\alpha_0-\x_{-1}^\mathrm{T}\al_{-1}-hy}{\alpha_1}+\v_{-1}^\mathrm{T}\x_{-1}\right)\\
&~~~~~~~~~~~\times \left(1-\beta^{*}_0-\beta^{*}_1\frac{1-\alpha_0-\x_{-1}^\mathrm{T}\al_{-1}-hy}{\alpha_1}-\x_{-1}^\mathrm{T}\be^{*}_{-1}\right)\\
&~~~~~~~~~~~\times f\left(\frac{1-\alpha_0-\x_{-1}^\mathrm{T}\al_{-1}-hy}{\alpha_1}|\x_{-1}\right)H'(y)dy\\
=&\frac{\beta^{*}_1}{\alpha_1}\int_{\mathbb{R}}\left(v_0+v_1\frac{1-\alpha_0-\x_{-1}^\mathrm{T}\al_{-1}-hy}{\alpha_1}+\v_{-1}^\mathrm{T}\x_{-1}\right)\Delta(\al,\be^{*},\x_{-1},y)\\
&\times f\left(\frac{1-\alpha_0-\x_{-1}^\mathrm{T}\al_{-1}-hy}{\alpha_1}|\x_{-1}\right)H'(y)dy\\
=&\int_{\mathbb{R}}\left(v_0+v_1\frac{1-\alpha_0-\x_{-1}^\mathrm{T}\al_{-1}-hy}{\alpha_1}+\v_{-1}^\mathrm{T}\x_{-1}\right)\Delta(\al,\be^{*},\x_{-1},y)\\
&\times f\left(\frac{1-\beta^{*}_0-\x_{-1}^\mathrm{T}\be^{*}_{-1}}{\beta^{*}_1}|\x_{-1}\right)H'(y)dy\\
&+O(1)\int_{\mathbb{R}}\left|v_0+v_1\frac{1-\alpha_0-\x_{-1}^\mathrm{T}\al_{-1}-hy}{\alpha_1}+\v_{-1}^\mathrm{T}\x_{-1}\right|\Delta^2(\al,\be^{*},\x_{-1},y)|H'(y)|dy\\
&+O(1)\frac{|\beta^{*}_1-\alpha_1|}{\alpha_1}\int_{\mathbb{R}}\left|v_0+v_1\frac{1-\alpha_0-\x_{-1}^\mathrm{T}\al_{-1}-hy}{\alpha_1}+\v_{-1}^\mathrm{T}\x_{-1}\right|\Delta(\al,\be^{*},\x_{-1},y)|H'(y)|dy\\
=&\int_{\mathbb{R}}\left(v_0+v_1\frac{1-\beta^{*}_0-\x_{-1}^\mathrm{T}\be^{*}_{-1}}{\beta_1}+\v_{-1}^\mathrm{T}\x_{-1}\right)\Delta(\al,\be^{*},\x_{-1},y)f\left(\frac{1-\beta^{*}_0-\x_{-1}^\mathrm{T}\be^{*}_{-1}}{\beta^{*}_1}|\x_{-1}\right)H'(y)dy\\
&+O(1)\int_{\mathbb{R}}\left|v_0+v_1\frac{1-\alpha_0-\x_{-1}^\mathrm{T}\al_{-1}-hy}{\alpha_1}+\v_{-1}^\mathrm{T}\x_{-1}\right|\Delta^2(\al,\be^{*},\x_{-1},y)|H'(y)|dy\\
&+O(1)\frac{|\beta^{*}_1-\alpha_1|}{\alpha_1}\int_{\mathbb{R}}\left|v_0+v_1\frac{1-\alpha_0-\x_{-1}^\mathrm{T}\al_{-1}-hy}{\alpha_1}+\v_{-1}^\mathrm{T}\x_{-1}\right|\Delta(\al,\be^{*},\x_{-1},y)|H'(y)|dy\\
&+O(1)\int_{\mathbb{R}}\Delta^2(\al,\be^{*},\x_{-1},y)|H'(y)|dy.
\end{align*}
\endgroup
Note that

\begin{equation*}
|\Delta(\al,\be^{*},\x_{-1},y)|\le C\left(h+|\x_{-1}^\mathrm{T}(\al_{-1}-\be^{*}_{-1})|+|1-\beta^{*}_0-\x_{-1}^\mathrm{T}\be^{*}_{-1}||\alpha_1-\beta^{*}_1|+|\alpha_0-\beta^{*}_0|\right).
\end{equation*}
So we have
\begin{equation*}
\begin{aligned}
&\int_{\mathbb{R}^p}(v_0+\v^\mathrm{T}\x)H\left(\frac{1-\alpha_0-\x^\mathrm{T}\al}{h}\right)f(\x)d\x\\
&~~~~~~~~~~~+\int_{\mathbb{R}^p}(v_0+\v^\mathrm{T}\x)\frac{1-\beta^{*}_0-\x^\mathrm{T}\be^{*}}{h}H'\left(\frac{1-\alpha_0-\x^\mathrm{T}\al}{h}\right)f(\x)d\x\\
=&-v_1\int_{\mathbb{R}^{p-1}}G\left(\frac{1-\beta^{*}_0-\x_{-1}^\mathrm{T}\be^{*}_{-1}}{\beta^{*}_1}|\x_{-1}\right)f_{-1}(\x_{-1})d\x_{-1}\\
&-\int_{\mathbb{R}^{p-1}}(v_0+\v_{-1}^\mathrm{T}\x_{-1})F\left(\frac{1-\beta^{*}_0-\x_{-1}^\mathrm{T}\be^{*}_{-1}}{\beta^{*}_1}|\x_{-1}\right)f_{-1}(\x_{-1})d\x_{-1}+O(h^2+\lVert\al-\be^{*}\rVert_2^2).
\end{aligned}
\end{equation*}
Note that
\begin{equation*}
v_1\int_{\mathbb{R}^{p-1}}G\left(\frac{1-\beta^{*}_0-\x_{-1}^\mathrm{T}\be^{*}_{-1}}{\beta^{*}_1}|\x_{-1}\right)f_{-1}(\x_{-1})d\x_{-1}=\mathbb{E}[v_1YX_1I\{1-Y\tX ^\mathrm{T}\tb^{*} \ge0\}|Y=1],
\end{equation*}
and
\begin{equation*}
\begin{aligned}
&\int_{\mathbb{R}^{p-1}}(v_0+\v_{-1}^\mathrm{T}\x_{-1})F\left(\frac{1-\beta^{*}_0-\x_{-1}^\mathrm{T}\be^{*}_{-1}}{\beta^{*}_1}|\x_{-1}\right)f_{-1}(\x_{-1})d\x_{-1}\\
=&\mathbb{E}[Y(v_0+\v_{-1}^\mathrm{T}\X_{-1})I\{1-Y\tX ^\mathrm{T}\tb^{*} \ge0\}|Y=1].
\end{aligned}
\end{equation*}
So
\begin{equation*}
\begin{aligned}
&\mathbb{E}\left\{Y\tilde{\v}^\mathrm{T}\tX \left(H\left(\frac{1-Y\tX ^\mathrm{T}\ta }{h}\right)+\frac{1-Y\tX ^\mathrm{T}\tb^{*} }{h}H'\left(\frac{1-Y\tX ^\mathrm{T}\ta }{h}\right)\right)\right\}\\
=&-\mathbb{E}[\tilde{\v}^\mathrm{T}Y\tX I\{1-Y\tX ^\mathrm{T}\tb^{*} \ge0\}]+O(h^2+\lVert\ta -\tb^{*} \rVert_2^2)\\
=&O(h^2+\lVert\ta -\tb^{*} \rVert_2^2).
\end{aligned}
\end{equation*}
\hfill\BlackBox

\begin{lemma}\label{lem:2}
	Suppose that conditions (C0)-(C4) hold. For any $\tilde{\v}\in\mathbb{R}^{p+1}$ with $\lVert\tilde{\v}\rVert_2=1$, we have
	\begin{equation*}
	\mathbb{E}\left\{\frac{1}{h}(\tilde{\v}^\mathrm{T}\tX )^2H'\left(\frac{1-Y\tX ^\mathrm{T}\ta }{h}\right)\right\}=\tilde{\v}^\mathrm{T}\mathbb{E}\left[\delta(1-Y\tX ^\mathrm{T}\tb^{*} )\tX \tX ^\mathrm{T}\right]\tilde{\v}+O(h+\lVert\ta -\tb^{*} \rVert_2),
	\end{equation*}
	uniformly in $\lVert\ta -\tb^{*} \rVert_2\le a_n$ with any $a_n\rightarrow0$.
\end{lemma}

\noindent
\textbf{Proof of Lemma \ref{lem:2}.}
Without loss of generality, assume that $\beta^{*}_1\ge c$. Then $\alpha_1\ge c/2$. For any $\v\in\mathbb{R}^p$,
\begin{equation*}
\begin{aligned}
&\mathbb{E}\left\{\frac{1}{h}(\tilde{\v}^\mathrm{T}\tX )^2H'\left(\frac{1-Y\tX ^\mathrm{T}\ta }{h}\right)\right\}\\
=&\pi_+\int_{\mathbb{R}^p}\frac{1}{h}(\tilde{\v}^\mathrm{T}\tilde{\x} )^2H'\left(\frac{1-\tilde{\x} ^\mathrm{T}\ta }{h}\right)f(\x)d\x+\pi_-\int_{\mathbb{R}^p}\frac{1}{h}(\tilde{\v}^\mathrm{T}\tilde{\x} )^2H'\left(\frac{1+\tilde{\x} ^\mathrm{T}\ta }{h}\right)g(\x)d\x.
\end{aligned}
\end{equation*}
We have
\begin{equation*}
\begin{aligned}
&\frac{1}{h}\int_{\mathbb{R}^p}(v_0+\v^\mathrm{T}\x)^2H'\left(\frac{1-\alpha_0-\x^\mathrm{T}\al}{h}\right)f(\x)d\x\\
=&\frac{1}{h}\int_{\mathbb{R}^{p-1}}\int_{\mathbb{R}}(v_0+v_1x_1+\v_{-1}^\mathrm{T}\x_{-1})^2H'\left(\frac{1-\alpha_0-x_1\alpha_1-\x_{-1}^\mathrm{T}\al_{-1}}{h}\right)f(x_1,\x_{-1})dx_1d\x_{-1}\\
=&-\frac{1}{\alpha_1}\int_{\mathbb{R}^{p-1}}f_{-1}(\x_{-1})\int_{-1}^1(v_0+v_1\frac{1-\alpha_0-\x_{-1}^\mathrm{T}\al_{-1}-hy}{\alpha_1}+\v_{-1}^\mathrm{T}\x_{-1})^2\\
&\times f\left(\frac{1-\alpha_0-\x_{-1}^\mathrm{T}\al_{-1}-hy}{\alpha_1}|\x_{-1}\right)H'(y)dyd\x_{-1}.\\
\end{aligned}
\end{equation*}
Note that
\begin{equation*}
\begin{aligned}
&\int_{-1}^1(v_0+\v_{-1}^\mathrm{T}\x_{-1})^2f\left(\frac{1-\alpha_0-\x_{-1}^\mathrm{T}\al_{-1}-hy}{\alpha_1}|\x_{-1}\right)H'(y)dy\\
=&\int_{-1}^1(v_0+\v_{-1}^\mathrm{T}\x_{-1})^2f\left(\frac{1-\beta^{*}_0-\x_{-1}^\mathrm{T}\be^{*}_{-1}}{\beta^{*}_1}|\x_{-1}\right)H'(y)dy\\&+O(1)\int_{-1}^1(v_0+\v_{-1}^\mathrm{T}\x_{-1})^2\Delta(\al,\be^{*},\x_{-1},y)|H'(y)|dy.
\end{aligned}
\end{equation*}
According to Condition (C2), we have
\begin{equation*}
\begin{aligned}
&\int_{-1}^12(v_0+\v_{-1}^\mathrm{T}\x_{-1})v_1\frac{1-\alpha_0-\x_{-1}^\mathrm{T}\al_{-1}-hy}{\alpha_1}f\left(\frac{1-\alpha_0-\x_{-1}^\mathrm{T}\al_{-1}-hy}{\alpha_1}|\x_{-1}\right)H'(y)dy\\
=&\int_{-1}^12(v_0+\v_{-1}^\mathrm{T}\x_{-1})v_1\frac{1-\beta^{*}_0-\x_{-1}^\mathrm{T}\be^{*}_{-1}}{\beta^{*}_1}f\left(\frac{1-\beta^{*}_0-\x_{-1}^\mathrm{T}\be^{*}_{-1}}{\beta^{*}_1}|\x_{-1}\right)H'(y)dy\\
&+O(1)\int_{-1}^12(v_0+\v_{-1}^\mathrm{T}\x_{-1})v_1\Delta(\al,\be^{*},\x_{-1},y)|H'(y)|dy,
\end{aligned}
\end{equation*}
and
\begin{equation*}
\begin{aligned}
&\int_{-1}^1v_1^2\left(\frac{1-\alpha_0-\x_{-1}^\mathrm{T}\al_{-1}-hy}{\alpha_1}\right)^2f\left(\frac{1-\alpha_0-\x_{-1}^\mathrm{T}\al_{-1}-hy}{\alpha_1}|\x_{-1}\right)H'(y)dy\\
=&\int_{-1}^1v_1^2\left(\frac{1-\beta^{*}_0-\x_{-1}^\mathrm{T}\be^{*}_{-1}}{\beta^{*}_1}\right)^2f\left(\frac{1-\beta^{*}_0-\x_{-1}^\mathrm{T}\be^{*}_{-1}}{\beta^{*}_1}|\x_{-1}\right)H'(y)dy\\&+O(1)\int_{-1}^1v_1^2\Delta(\al,\be^{*},\x_{-1},y)|H'(y)|dy.
\end{aligned}
\end{equation*}
Therefore,
\begin{equation*}
\begin{aligned}
&\frac{1}{h}\int_{\mathbb{R}^p}(v_0+\v^\mathrm{T}\x)^2H'\left(\frac{1-\alpha_0-\x^\mathrm{T}\al}{h}\right)f(\x)d\x\\
=&-\frac{1}{\beta^{*}_1}\int_{\mathbb{R}^{p-1}}f_{-1}(\x_{-1})\int_{-1}^1(v_0+v_1\frac{1-\beta^{*}_0-\x_{-1}^\mathrm{T}\be^{*}_{-1}}{\beta^{*}_1}+\v_{-1}^\mathrm{T}\x_{-1})^2\\
&\times f\left(\frac{1-\beta^{*}_0-\x_{-1}^\mathrm{T}\be^{*}_{-1}}{\beta^{*}_1}|\x_{-1}\right)H'(y)dyd\x_{-1}\\
&+O(1)\int_{\mathbb{R}^{p-1}}f_{-1}(\x_{-1})\int_{-1}^1(v_0+v_1+\v_{-1}^\mathrm{T}\x_{-1})^2\Delta(\al,\be^{*},\x_{-1},y)|H'(y)|dyd\x_{-1}\\
&+O(1)\frac{|\alpha_1-\beta^{*}_1|}{\alpha_1\beta^{*}_1}\int_{\mathbb{R}^{p-1}}f_{-1}(\x_{-1})\int_{-1}^1(v_0+v_1\frac{1-\beta^{*}_0-\x_{-1}^\mathrm{T}\be^{*}_{-1}}{\beta^{*}_1}+\v_{-1}^\mathrm{T}\x_{-1})^2|H'(y)|dyd\x_{-1}\\
=&\tilde{\v}^\mathrm{T}\mathbb{E}\left[\delta(1-Y\tX ^\mathrm{T}\tb^{*} )\tX \tX ^\mathrm{T}|Y=1\right]\tilde{\v}+O(h+\lVert\ta -\tb^{*} \rVert_2).
\end{aligned}
\end{equation*}
Then we get
\begin{equation*}
\mathbb{E}\left\{\frac{1}{h}(\tilde{\v}^\mathrm{T}\tX )^2H'\left(\frac{1-Y\tX ^\mathrm{T}\ta }{h}\right)\right\}=\tilde{\v}^\mathrm{T}\mathbb{E}\left[\delta(1-Y\tX ^\mathrm{T}\tb^{*} )\tX \tX ^\mathrm{T}\right]\tilde{\v}+O(h+\lVert\ta -\tb^{*} \rVert_2).
\end{equation*}
\hfill\BlackBox

Define $K(\tX,\tth)=|1+\tX^{\mathrm{T}}\tth|$ and
\begin{eqnarray*}
	H_h(\alp)=H\left(\frac{1-Y\tX ^\mathrm{T}\alp}{h}\right)-I\{\epsilon\ge0\}+\frac{\epsilon}{h}H'\left(\frac{1-Y\tX ^\mathrm{T}\alp}{h}\right).
\end{eqnarray*}

\begin{lemma}\label{lem:3}
	Suppose that conditions (C0)-(C4) hold. For  some $t>0$ and any $\tilde{\v},\tth\in\mathbb{R}^{p+1}$ with $\lVert\tilde{\v}\rVert_2=1$ and $\|\tth\|_{2}=1$, we have
	\begin{equation*}
	\begin{aligned}
	\mathbb{E}\{\tilde{\v}^\mathrm{T}Y\tX  H_h(\alp)\}^2\exp(t|\tilde{\v}^\mathrm{T}\tX| K(\tX,\tth))
	=O(h+\lVert\alp-\tb^{*} \rVert_2+\lVert\alp-\tb^{*} \rVert_2^2/h),
	\end{aligned}
	\end{equation*}
	uniformly in $\lVert\ta -\tb^{*} \rVert_2\le a_n$ with any $a_n\rightarrow0$.
\end{lemma}

\noindent
\textbf{Proof of Lemma \ref{lem:3}.}
We have
\begin{equation*}
\begin{aligned}
&\int_{\mathbb{R}^p}(\tilde{\v}^\mathrm{T}\tilde{\x} )^2\exp(t|\tilde{\v}^\mathrm{T}\tilde{\x}| K(\tilde{\x},\tth))\left[H\left(\frac{1-\tilde{\x} ^\mathrm{T}\alp}{h}\right)-I\{\epsilon\ge0\}\right]^2f(\x)d\x\\
=&-\frac{h}{\alpha_1}\int_{\mathbb{R}^{p-1}}f_{-1}(\x_{-1})\int_{\mathbb{R}}(\tilde{\v}^\mathrm{T}\tilde{\x}^{*} )^2\exp(t|\tilde{\v}^\mathrm{T}\tilde{\x}^{*}| K(\tilde{\x}^{*},\tth))\\
&~~~~~~~~~~\times\left[H(y)-I\left\{1-\beta^{*}_0-\beta^{*}_1\frac{1-\alpha_0-\x_{-1}^\mathrm{T}\al_{-1}-hy}{\alpha_1}-\x_{-1}^\mathrm{T}\be^{*}_{-1}\ge0\right\}\right]^2\\
&~~~~~~~~~~\times f\left(\frac{1-\alpha_0-\x_{-1}^\mathrm{T}\al_{-1}-hy}{\alpha_1}|\x_{-1}\right)dyd\x_{-1},
\end{aligned}
\end{equation*}
where $I\left\{1-\beta^{*}_0-\beta^{*}_1\frac{1-\alpha_0-\x_{-1}^\mathrm{T}\al_{-1}-hy}{\alpha_1}-\x_{-1}^\mathrm{T}\be^{*}_{-1}\ge0\right\}=I\{y\ge\frac{\alpha_1}{h}\Delta(\al,\be^{*},\x_{-1},0)\}$ and $\tilde{\x}^{*}$ denotes $\tilde{\x}$ with $x_{1}$ being replaced by $\frac{1-\alpha_0-\x_{-1}^\mathrm{T}\al_{-1}-hy}{\alpha_1}$. According to Condition (C2), (C3) and (C4), note that
\begin{equation*}
\begin{aligned}
&\int_{\mathbb{R}}(\tilde{\v}^\mathrm{T}\tilde{\x}^{*} )^2\exp(t|\tilde{\v}^\mathrm{T}\tilde{\x}^{*}| K(\tilde{\x}^{*},\tth))\left[H(y)-I\{y\ge\frac{\alpha_1}{h}\Delta(\al,\be^{*},\x_{-1},0)\}\right]^2\\
&\times f\left(\frac{1-\alpha_0-\x_{-1}^\mathrm{T}\al_{-1}-hy}{\alpha_1}|\x_{-1}\right)dy\\
\le&C\left(\v_{-1}^\mathrm{T}\x_{-1}\right)^2\exp(t^{'}|\v^{\mathrm{T}}_{-1}\x_{-1}|^{2}+t^{'}(1+|\tee_{-1}^\mathrm{T}\x_{-1}|+
|\al_{-1}^\mathrm{T}\x_{-1}|+|\be_{-1}^{*\mathrm{T}}\x_{-1}|)^{2})\\
&\times\left(1+\left|\frac{\alpha_1}{h}\Delta(\al,\be^{*},\x_{-1},0)\right|\right),
\end{aligned}
\end{equation*}
for some $t^{'}>0$.
Therefore, by (C3)
\begin{equation*}
\int_{\mathbb{R}^p}(\tilde{\v}^\mathrm{T}\tilde{\x} )^2\exp(t|\tilde{\v}^\mathrm{T}\tilde{\x}| K(\tilde{\x},\tth))\left[H\left(\frac{1-\tilde{\x}^\mathrm{T}\alp}{h}\right)-I\{\epsilon\ge0\}\right]^2f(\x)d\x\le O(h+\lVert\alp-\tb^{*} \rVert_2).
\end{equation*}
On the other hand, we can easily prove that
\begin{equation*}
\begin{aligned}
&\int_{\mathbb{R}^p}(\tilde{\v}^\mathrm{T}\tilde{\x} )^2\exp(t|\tilde{\v}^\mathrm{T}\tilde{\x}| K(\tilde{\x},\tth))\left[\frac{\epsilon}{h}H'\left(\frac{1-\tilde{\x} ^\mathrm{T}\alp}{h}\right)\right]^2f(\x)d\x\\
=&-\frac{h\beta_1^{*2}}{\alpha_1}\int_{\mathbb{R}^{p-1}}f_{-1}(\x_{-1})\int_{-1}^1(\tilde{\v}^\mathrm{T}\tilde{\x}^{*} )^2\exp(t|\tilde{\v}^\mathrm{T}\tilde{\x}^{*}| K(\tilde{\x}^{*},\tth))\left[\frac{\Delta(\al,\be^{*},\x_{-1},y)}{h}H'(y)\right]^2\\
&~~~~~~~~~~\times f\left(\frac{1-\alpha_0-\x_{-1}^\mathrm{T}\al_{-1}-hy}{\alpha_1}|\x_{-1}\right)dyd\x_{-1}\\
\le& O(h+\lVert\alp-\tb^{*} \rVert_2^2/h).
\end{aligned}
\end{equation*}
Now we complete the proof of the lemma.
\hfill\BlackBox

\subsection{Proofs of the Main Results}
\label{app:main}
After introducing and proving the above three lemmas, we begin to prove Proposition \ref{prop:C} and \ref{prop:D}, Theorem \ref{thm:bahadur} and \ref{thm:asym}.

\noindent
\textbf{Proof of Proposition \ref{prop:C}.}
Recall that $\epsilon_i=1-y_i\tX _i^\mathrm{T}\tb^{*} $. Define $\Delta(\alp)=\alp-\tb^{*}$, and
\begin{equation*}
\begin{aligned}
\C_{n,h}(\alp)&=\A_{n,h}(\alp)-\frac{1}{n}\sum_{i=1}^ny_i\tX _iI\{\epsilon_i\ge0\}\\
&=\frac{1}{n}\sum_{i=1}^n y_i\tX _i\left[H\left(\frac{1-y_i\tX _i^\mathrm{T}\ta }{h}\right)-I\{\epsilon_i\ge0\}+\frac{1-y_i\tX _i^\mathrm{T}\tb^{*} }{h}H'\left(\frac{1-y_i\tX _i^\mathrm{T}\ta }{h}\right)\right].
\end{aligned}
\end{equation*}
Let $\C_{n,h}=\C_{n,h}(\tb^{*})$. Note that $\lVert \C_{n,h}\rVert_2=\sup_{\tilde{\v}\in\mathbb{R}^{(p+1)},\lVert\tilde{\v}\rVert_2=1}|\tilde{\v}^\mathrm{T}\C_{n,h}|$.

Let $S_{1/2}^p$ be a 1/2 net of the unit sphere $S^p$ in the Euclidean distance in $\mathbb{R}^{p+1}$. According to the proof of Lemma 3 in \cite{cai2010optimal}, we have $d_{p+1}$:=Card$(S_{1/2}^p)\le 5^{p+1}$. Let $\tilde{\v}_1,...,\tilde{\v}_{d_{p+1}}$ be the centers of the $d_{p+1}$ elements in the net. Therefore for any $\tilde{\v}$ in $S^p$, we have $\lVert \tilde{\v}-\tilde{\v}_j\rVert_2\le 1/2$ for some $j$. Therefore, $\lVert \C_{n,h}\rVert_2\le 2\sup_{j\le d_{p+1}}|\tilde{\v}_j^\mathrm{T}\C_{n,h}|$.
It is easy to see that for any $M>0$, there exists a set of points in $\R^{p+1}$, $\{\alp_{k}, 1\leq k\leq s_{p+1}\}$ with $s_{p+1}\leq n^{M(p+1)}$, such that for any $\alp$ in the ball $\|\alp-\tb^{*}\|_2\leq a_{n}$, we have $\|\alp-\alp_{k}\|_2\leq 2\sqrt{p+1}a_{n}/n^{M}$ for some $1\leq k\leq s_{p+1}$ and $\|\alp_{k}-\tb^{*}\|_2\leq a_{n}$.

Define
\begin{equation*}
\begin{aligned}
C_{n,h,j}(\alp)&=\frac{1}{n}\sum_{i=1}^n \tilde{\v}_j^\mathrm{T}y_i\tX _i\left[H\left(\frac{1-y_i\tX _i^\mathrm{T}\ta }{h}\right)-I\{\epsilon_i\ge0\}+\frac{\epsilon_i}{h}H'\left(\frac{1-y_i\tX _i^\mathrm{T}\ta }{h}\right)\right]\\&\triangleq\frac{1}{n}\sum_{i=1}^n \tilde{\v}_j^\mathrm{T}y_i\tX _iH_{h,i}(\alp).
\end{aligned}
\end{equation*}
According to the proof of Proposition 4.1 in \cite{chen2018quantile},  it is enough to show that $$\sup_j\sup_k|{C}_{n,h,j}(\alp_k)|=O_\mathbb{P}\left(\sqrt{\frac{ph\log n}{n}}+a_n^2+h^2\right).$$
Since $H$ and $xH'(x)$ are bounded, it is easy to see that
\begin{eqnarray*}
	|H_{h,i}(\alp)|\leq C(1+|\tX^{\mathrm{T}}_{i}(\alp-\tb^{*})|/\|\alp-\tb^{*}\|_{2})=:K(\tX_{i},\alp,\tb^{*}).
\end{eqnarray*}
By Lemma \ref{lem:3}, we have for some $t>0$,
\begin{equation*}
\mathbb{E}(\tilde{\v}_j^\mathrm{T}y_i\tX _iH_{h,i}(\alp))^2\exp(t|\tilde{\v}_j^\mathrm{T}\tX _i|K(\tX_{i},\alp,\tb^{*}))\le Ch(1+\lVert\alp-\tb^{*} \rVert_2/h+\lVert\alp-\tb^{*} \rVert_2^2/h^2).
\end{equation*}
By  $\sqrt{p\log n}=o(\sqrt{nh})$ and Lemma 1 in \cite{cailiu2011}, we can get for any $\gamma>0$, there exists a constant $C$ such that
\begin{equation*}
\sup\limits_{j}\sup\limits_{k}\mathbb{P}\left(|C_{n,h,j}(\alp_k)-\mathbb{E}C_{n,h,j}(\alp_k)|\ge C\sqrt{\frac{ph\log n}{n}}\right)=O(n^{-\gamma p}).
\end{equation*}
The remaining work is to give a bound for $\mathbb{E}C_{n,h,j}(\alp_k)$. According to Lemma \ref{lem:1}, we know that $\mathbb{E}C_{n,h,j}(\alp_k)=O(h^2+\lVert\alp_k-\tb^{*} \rVert_2^2)$. Hence, $\sup_j\sup_k|\mathbb{E}C_{n,h,j}(\alp_k)|=O(h^2+\lVert\alp_k-\tb^{*} \rVert_2^2)$.
Combining with the above analysis, the proof is completed.
\hfill\BlackBox
\\

\noindent
\textbf{Proof of Proposition \ref{prop:D}.}
For simplicity, denote $\D(\tb^{*} )$ by $\D$. According to the proof of Lemma 3 in \cite{cai2010optimal}, for $\D_{n,h}$ we have
\begin{equation*}
\lVert \D_{n,h}-\D\rVert\le 10\sup\limits_{j\le b_{p+1}}|\tilde{\v}_j^\mathrm{T}(\D_{n,h}-\D)\tilde{\v}_j|.
\end{equation*}
where $\tilde{\v}_j$, $1\le j\le b_{p+1}$, are some non-random vectors with $\lVert \tilde{\v}_j\rVert_2=1$ and $b_{p+1}\le 5^{p+1}$. Define
\begin{equation*}
D_{n,h,j}(\alp)=\frac{1}{nh}\sum_{i=1}^n (\tilde{\v}_j^\mathrm{T}\tX _i)^2H'\left(\frac{1-y_i\tX _i^\mathrm{T}\alp}{h}\right).
\end{equation*}
When $\lVert\tb _0-\tb^{*} \rVert_2\le a_n$, then
\begin{equation*}
\sup\limits_{j\le b_{p+1}}|\tilde{\v}_j^\mathrm{T}(\D_{n,h}-\D)\tilde{\v}_j|\le\sup\limits_{j\le b_{p+1}}\sup\limits_{\lVert\ta -\tb^* \rVert_2\le a_n}|D_{n,h,j}(\alp)-\tilde{\v}_j^\mathrm{T}\D\tilde{\v}_j|.
\end{equation*}
As the proof of Lemma \ref{lem:3}, we obtain that
\begin{equation*}
\mathbb{E}\left[\tilde{\v}^\mathrm{T}\widetilde{\x}_iH'\left(\frac{{\epsilon}_i-y_i\widetilde{\x}_i^\mathrm{T}\Delta(\alp)}{h}\right)\right]^2=O(h).
\end{equation*}
According to the proof of Proposition 4.2 in \cite{chen2018quantile}, $D_{n,h,j}$ satisfies
\begin{equation*}
\sup\limits_{j}\sup\limits_{k}|D_{n,h,j}(\alp_k)-\mathbb{E}D_{n,h,j}(\alp_k)|=O_\mathbb{P}\left(\sqrt{\frac{p\log n}{nh}}\right).
\end{equation*}
The remaining work is to give a bound for $\mathbb{E}D_{n,h,j}(\alp_k)-\tilde{\v}_j^\mathrm{T}\D\tilde{\v}_j$. From Lemma \ref{lem:2}, we obtain that
\begin{equation*}
\mathbb{E}D_{n,h,j}(\alp)-\tilde{\v}_j^\mathrm{T}\D\tilde{\v}_j=O(h+\lVert\alp-\tb^{*} \rVert_2).
\end{equation*}
Hence, $\sup_j\sup_k|\mathbb{E}D_{n,h,j}(\alp_k)-\tilde{\v}_j^\mathrm{T}\D\tilde{\v}_j|=O(h+\lVert\alp_k-\tb^{*} \rVert_2)$. Combining with the above analysis, the proof of the proposition is completed.
\hfill\BlackBox
\\

\noindent
\textbf{Proof of Theorem \ref{thm:bahadur} and \ref{thm:asym}.} We first assume that  $\|\tb_{0}-\tb^{*}\|=O_{\mathbb{P}}(a_{n})$ with $a_{n}=o(1)$ and $a_{n}=O(h)$.
For independent random vectors $\{(y_i,\tX _i),i=1,...,n\}$ with $\sup_{j}\mathbb{E}|X_{j}|^3=O(1)$, we can see that
\begin{equation*}
\lVert\frac{1}{n}\sum_{i=1}^n y_i\tX _iI(\epsilon_i\ge0)\rVert_2=O_\mathbb{P}(\sqrt{p/n}).
\end{equation*}
Note that $\|\tb_{0}-\tb^{*}\|=o_{\mathbb{P}}(1)$ and $\|\tb^{*}\|_{2}\leq C$.
From the above result and Proposition \ref{prop:C} and \ref{prop:D} we know that for the estimator $\tb $ and the true parameter $\tb ^*$,
\begin{equation*}
\tb -\tb ^*=\D(\tb ^*)^{-1}\left(\frac{1}{n}\sum_{i=1}^ny_i\tX _iI\{\epsilon_i\ge0\}-\lambda \binom{0}{\be^{*}}+\lambda \binom{0}{\be^{*}-\be_0}\right)+\r_n.
\end{equation*}
with
\begin{equation*}
\lVert \r_n\rVert_2=O_\mathbb{P}\left(\sqrt{\frac{p^2\log n}{n^2h}}+\sqrt{\frac{p\log n}{nh}}\lambda+\sqrt{\frac{ph\log n}{n}}+\lambda h+h^2\right).
\end{equation*}
Since $\lambda\leq h$ and $\|\tb_{0}-\tb^{*}\|=O_{\mathbb{P}}(a_{n})$, we have
\begin{equation*}
\tb -\tb ^*=\D(\tb ^*)^{-1}\left(\frac{1}{n}\sum_{i=1}^ny_i\tX _iI\{\epsilon_i\ge0\}-\lambda \binom{0}{\be^{*}}\right)+\r_n.
\end{equation*}
with
\begin{eqnarray}\label{cdse}
\lVert \r_n\rVert_2=O_\mathbb{P}\left(\sqrt{\frac{p^2\log n}{n^2h}}+\sqrt{\frac{ph\log n}{n}}+h^2\right).
\end{eqnarray}
Note that $h_g\ge\sqrt{p/n}$, then $\sqrt{\frac{p^2\log n}{n^2h_g}}\le\sqrt{\frac{ph_g\log n}{n}}$.

For $q=1$, it is easy to see that Theorem \ref{thm:bahadur} holds. Suppose the theorem holds for $q=g-1$ with $g\ge2$. Note that $p=O(m/(\log n)^2)$ and $\lambda=O(1/\log n)$, then $h_{g-1}=O(1/\log n)$ and we have $\sqrt{ph_{(g-1)}(\log n)/n}=O(\sqrt{p/n})$. Then we have $a_n=\max\{\lambda^{2},\sqrt{p/n},(p/m)^{2^{g-2}}\}=O(h_{g})$ for $q=g$ with initial estimator $\widehat{\beta}_0=\widehat{\beta}^{(g-1)}$. Now we complete the proof of Theorem \ref{thm:bahadur} by (\ref{cdse}). Theorem \ref{thm:asym} follows directly from Theorem \ref{thm:bahadur} and the Lindeberg-Feller central limit theorem.

\hfill\BlackBox

\noindent
\textbf{Proof of Theorem \ref{col:plugin}.} To prove Theorem \ref{col:plugin}, we first introduce the following lemma, which shows  that $\widehat{\G}(\tb^{(q)})$ is a consistent estimator of ${\G}(\tb^{*})$.

\begin{lemma}\label{le4} Under the conditions of Theorem 2 and $a_{n}\rightarrow 0$,  we have
	\begin{eqnarray*}
		\sup_{\|\alp-\tb^*\|_{2}\leq a_{n}}\|\widehat{\G}(\alp)-\G(\tb^*)\|=o_{\mathbb{P}}(1).
	\end{eqnarray*}
\end{lemma}

\noindent
\textbf{Proof of Lemma \ref{le4}}. Let $\tilde{\v}_j$, $1\le j\le b_{p+1}$ be defined as in the proof Proposition \ref{prop:D}. Define
\begin{equation*}
G_{n,j}(\alp)=\frac{1}{n}\sum_{i=1}^n (\tilde{\v}_j^\mathrm{T}\tX _i)^2I\{1-Y\tX_{i}^\mathrm{T}\alp\ge0\}.
\end{equation*}
By (C3) and Lemma 1 in \cite{cailiu2011},  we can show that,  for any $\gamma>0$, there exists a constant $C>0$ such that
\begin{eqnarray*}
	\max_{j}\sup_{\|\alp-\tb^*\|_{2}\leq a_{n}}\mathbb{P}\left(\Big{|}G_{n,j}(\alp)-\mathbb{E}G_{n,j}(\alp)\Big{|}\geq C\sqrt{\frac{p\log n}{n}}\right)=O(n^{-\gamma p}).
\end{eqnarray*}
Let $\alp_{k}$, $1\leq k\leq s_{p}$, be defined as in the proof of Proposition \ref{prop:C}.
Therefore
\begin{eqnarray}\label{asds0}
\max_{1\leq j\leq b_{p+1}}\max_{1\leq k\leq s_{p}}\Big{|}G_{n,j}(\alp_{k})-\mathbb{E}G_{n,j}(\alp_{k})\Big{|}=O_{\mathbb{P}}\Big{(}\sqrt{\frac{p\log n}{n}}\Big{)}.
\end{eqnarray}
Put $t_{M}=2\sqrt{p+1}a_{n}/n^{M}$.  In the following, we  show that
\begin{eqnarray}\label{asds1}
\mathbb{P}\left(\max_{j}\max_{k}\sup_{\|\alp-\alp_{k}\|_{2}\leq t_{M}}\Big{|}G_{n,j}(\alp)-G_{n,j}(\alp_{k})\Big{|}\geq C\sqrt{\frac{p\log n}{n}}\right)=o(1)
\end{eqnarray}
and
\begin{eqnarray}\label{asds2}
\max_{j}\max_{k}\sup_{\|\alp-\alp_{k}\|_{2}\leq t_{M}}\Big{|}\mathbb{E}G_{n,j}(\alp)-\mathbb{E}G_{n,j}(\alp_{k})\Big{|}\leq C\sqrt{\frac{p\log n}{n}}.
\end{eqnarray}
Note that for $\|\alp-\alp_{k}\|_{2}\leq t_{M}$,
\begin{eqnarray*}
	\Big{|}I\{1-y_{i}\tX_{i}^\mathrm{T}\alp\ge0\}-I\{1-y_{i}\tX_{i}^\mathrm{T}\alp_{k}\ge0\}\Big{|}&\leq& I\{-t_{M}\|\tX_{i}\|_{2}\leq 1-y_{i}\tX^{\mathrm{T}}_{i}\alp_{k}\leq
	t_{M}\|\tX_{i}\|_{2}\}\cr
	&\leq&I\{-t_{M}n\leq 1-y_{i}\tX^{\mathrm{T}}_{i}\alp_{k}\leq
	t_{M}n\}\cr
	& &+I\{\|\tX_{i}\|_{2}\geq n\}.
\end{eqnarray*}
By (C3), we have
\begin{eqnarray}\label{asds3}
\mathbb{P}(\max_{1\leq i\leq n}\|\tX_{i}\|_{2}\geq n)=o(1).
\end{eqnarray}
Define
\begin{eqnarray*}
	Z_{i,j,k}(\alp_{k})=(\tilde{\v}_j^\mathrm{T}\tX _i)^2I\{-t_{M}n\leq 1-y_{i}\tX^{\mathrm{T}}_{i}\alp_{k}\leq
	t_{M}n\}.
\end{eqnarray*}
By (C2) and (C3), we have
\begin{eqnarray}\label{asds4}
\mathbb{E}Z_{i,j,k}(\alp_{k})\leq \sqrt{\mathbb{E}(\tilde{\v}_j^\mathrm{T}\tX _i)^4}\sqrt{\mathbb{P}(-t_{M}n\leq 1-y_{i}\tX^{\mathrm{T}}_{i}\alp_{k}\leq
	t_{M}n)}
=O(\sqrt{t_{M}n})
\end{eqnarray}
and
\begin{eqnarray*}
	\mathbb{E}(Z_{i,j,k}(\alp_{k}))^{2}\exp(t_{0}Z_{i,j,k}(\alp_{k}))=O(\sqrt{t_{M}n}).
\end{eqnarray*}
Note that
\begin{eqnarray}\label{asds5}
\sup_{\|\alp-\alp_{k}\|_{2}\leq t_{M}}\Big{|}G_{n,j}(\alp)-G_{n,j}(\alp_{k})\Big{|}I\{\max_{1\leq i\leq n}\|\tX_{i}\|_{2}< n\}\leq \frac{1}{n}\sum_{i=1}^{n}Z_{i,j,k}(\alp_{k}).
\end{eqnarray}
By (C3) and Lemma 1 in \cite{cailiu2011}, for any $\gamma>0$, there exists a constant $C>0$ such that
\begin{eqnarray*}
	\max_{j,k}\mathbb{P}\left(\Big{|}\frac{1}{n}\sum_{i=1}^{n}(Z_{i,j,k}(\alp_{k})-\mathbb{E} Z_{i,j,k}(\alp_{k}))\Big{|}\geq C\sqrt{\frac{p\log n}{n}}\right)=O(n^{-\gamma p}).
\end{eqnarray*}
By (\ref{asds3})-(\ref{asds5}), we can see that (\ref{asds1}) and (\ref{asds2}) hold.

By (\ref{asds0}), (\ref{asds1}) and the definition of $\tilde{\v}_j$,  we have
\begin{eqnarray*}
	\sup_{\|\alp-\tb^*\|_{2}\leq a_{n}}\|\widehat{\G}(\alp)-\G(\alp)\|=O_{\mathbb{P}}\Big{(}\sqrt{\frac{p\log n}{n}}\Big{)}.
\end{eqnarray*}
Moreover, we have
\begin{eqnarray*}
	\Big{|}I\{1-y_{i}\tX_{i}^\mathrm{T}\alp\ge0\}-I\{1-y_{i}\tX_{i}^\mathrm{T}\tb^{*}\ge0\}\Big{|}&\leq&I\{-\sqrt{a_{n}}\leq 1-y_{i}\tX^{\mathrm{T}}_{i}\alp_{k}\leq
	\sqrt{a_{n}}\}\cr
	& &+I\{|\tX^{\mathrm{T}}_{i}(\alp-\tb^{*})|/\|\alp-\tb^{*}\|_{2}\geq a^{-1/2}_{n}\}.
\end{eqnarray*}
By this inequality, (C2) and (C3), it is easy to show that
\begin{eqnarray*}
	\mathbb{E}(\tilde{\v}^\mathrm{T}\tX _i)^2I\{1-Y\tX_{i}^\mathrm{T}\alp\ge0\}-\mathbb{E}(\tilde{\v}^\mathrm{T}\tX _i)^2I\{1-Y\tX_{i}^\mathrm{T}\tb^{*}\ge0\}=o(1)
\end{eqnarray*}
uniformly in $\|\v\|_{2}=1$ and $\|\alp-\tb^{*}\|_{2}\leq a_{n}$. This implies that $\sup_{\|\alp-\tb^{*}\|_{2}\leq a_{n}}\|\G(\alp)-\G(\tb^{*})\|=o_\mathbb{P}(1)$.
\hfill\BlackBox
\\

Now we prove Theorem \ref{col:plugin}. Without loss of generality, we can assume that $\|\tilde{\v}\|_{2}=1$.
By the consistency of $\widehat{\D}(\tb^{(q-1)})$ and $\widehat{\G}(\tb^{(q)})$ (see Lemma \ref{le4}), we have $$\hat{\sigma}_{n,q} \rightarrow \sqrt{\tilde{\v}^\mathrm{T}\D(\tb ^*)^{-1}\G(\tb ^*)\D(\tb ^*)^{-1}\tilde{\v}}$$ as $n,p\rightarrow \infty$. This implies the theorem.
\hfill\BlackBox
\\

\subsection{Proof of Auxiliary Results}
\label{app:aux}
\noindent
\textbf{Proof of Proposition \ref{prop:efficient}.}
By Proposition \ref{prop:D},  in the $g$-th iteration, we have
\begin{equation*}
\left\lVert N^{-1}\sum_{k=1}^{N}\V_{k}-\D(\tb ^*)\right\rVert=O_\mathbb{P}\left(\sqrt{\frac{p\log n}{nh_{g}}}+h_{g}\right).
\end{equation*}
Therefore, it suffices to show that
\begin{equation*}
\left\lVert \hat{\V}_{1}-\D(\tb ^*)\right\rVert=O_\mathbb{P}\left(m^{-\delta}\right)
\end{equation*}
for some $\delta>0$. With the notation in the proof of Proposition \ref{prop:D}, we have
\begin{equation*}
\sup\limits_{j}\sup\limits_{k}|D_{m,h,j}(\alp_k)-\mathbb{E}D_{m,h,j}(\alp_k)|=O_\mathbb{P}\left(\sqrt{\frac{p\log n}{mh}}\right)
\end{equation*}
with $h=\sqrt{p/m}$. Also,  $|\mathbb{E}D_{m,h,j}(\alp_k)-\tilde{\v}_j^\mathrm{T}\D\tilde{\v}_j|=O(h+\lVert\alp_k-\tb^{*} \rVert_2)$ uniformly in $j,k$. This completes the proof as $p=O(m^{\gamma})$ for some $0<\gamma<1$.
\hfill\BlackBox
\\

\noindent
\textbf{Proof of Claim \ref{claim:sim}.}
Let us define $\bar{\eps} = Y\eps = (\bar{\epsilon}^1,\ldots,\bar{\epsilon}^p)$. It is easy to show it follows normal distribution $\mathcal{N}(0,\sigma^2 \textbf{I})$. By the construction of $\X$ (i.e., $\X = Y\textbf{1}+\eps$), we have $Y\X = Y^2\textbf{1} + Y\eps = \textbf{1} + \bar{\eps}$ and $Y\tX = \binom{Y}{\textbf{1}+\bar{\eps}}$. Recall that $S(\tb^{*}) = -\mathbb{E}[I\{1-Y\tX^\mathrm{T}\tb^{*}\geq 0\}Y\tX]$. Therefore we have
\[
S(\tb^*) =-\mathbb{E}\left[I\left\{1-\frac{1}{a}(\textbf{1}+\bar{\eps})^\mathrm{T}\textbf{1}\geq 0\right\}\binom{Y}{\textbf{1}+\bar{\eps}}\right] = -\mathbb{E}\left[I\{a\geq p + \textbf{1}^\mathrm{T} \bar{\eps}\}\binom{Y}{\textbf{1}+\bar{\eps}}\right].
\]
In order to show that $S(\tb^*) = 0$, we only need to show that $\mathbb{E}\left[I\{a\geq p + \textbf{1}^\mathrm{T} \bar{\eps}\}Y\right] =0 $ and $\mathbb{E}\left[I\{a\geq p + \textbf{1}^\mathrm{T} \bar{\eps}\}(1+\bar{\epsilon}^i)\right] = 0$ for $i=1,\ldots,p$. The first equation holds because $Y$ is independent of $\bar{\eps}$ and $\mathbb{E}[Y] = p_+-p_- = 0$. To show the second equation, we note that for any $i,j \in\{1,\ldots,p\}$, we have
$\mathbb{E}[I\{a\geq p + \textbf{1}^\mathrm{T} \bar{\eps}\}(1+\bar{\epsilon}^j)] = \mathbb{E}[I\{a\geq p + \textbf{1}^\mathrm{T} \bar{\eps}\}(1+\bar{\epsilon}^i)]$ by distributional symmetry of $\bar{\epsilon}^i$ and $\bar{\epsilon}^j$. Therefore it is enough to show that
\[
\sum_{j=1}^{p}\mathbb{E}[I\{a\geq p + \textbf{1}^\mathrm{T} \bar{\eps}\}(1+\bar{\epsilon}^j)] = \mathbb{E}[I\{ \nu\leq a \}\nu]=0,
\]
where $\nu = p+\textbf{1}^\mathrm{T} \bar{\eps}$. Recall that $a$ satisfies $\int_{-\infty}^{a} \phi_1(x)x dx=0$ where $\phi_1(x)$ is the p.d.f. of the distribution $\mathcal{N}(p,\sigma^2 p)$. Since $\nu$ follows the normal distribution $\mathcal{N}(p,\sigma^2 p)$, we have $E[I\{ \nu\leq a \}\nu] = 0$. Therefore we have shown that $S(\tb^*)=0$. By the convexity of the loss  function and uniqueness of the minimizer, we have proved that $\tb^*$ is the true coefficient under the given setting.
\hfill\BlackBox

\bibliographystyle{chicago}
	\bibliography{refs}
	
\end{document}